%% file: sample-sigconf.tex
\documentclass[sigconf]{acmart}

\input{math_commands.tex}

\usepackage{annote-equations} 

\usepackage[utf8]{inputenc} 
\usepackage[T1]{fontenc}    
\usepackage{hyperref}       
\usepackage{url}            
\usepackage{booktabs}       
\usepackage{amsfonts}       
\usepackage{nicefrac}       
\usepackage{microtype}      
\usepackage{xcolor}         
\usepackage{xspace}
\usepackage{bm} 
\usepackage{multirow}
\usepackage{array}
\usepackage{colortbl}
\usepackage{amsfonts}
\usepackage{amsmath}
\usepackage{amsthm}
\usepackage{subfigure}
\usepackage{listings}
\usepackage{threeparttable}
\usepackage{wrapfig}
\usepackage{algorithm}
\usepackage{algpseudocode}
\usepackage{diagbox}
\usepackage{graphicx}
\usepackage{diagbox}
\usepackage{pifont}
\usepackage{booktabs}
\usepackage{float}
\usepackage[title]{appendix}
\usepackage{fontawesome5}
\usepackage{bbding}
\usepackage{makecell}
\usepackage[most]{tcolorbox}
\usepackage{color, colortbl}
\usepackage{arydshln}

\usepackage{tikz}
\newcommand*\circled[1]{\tikz[baseline=(char.base)]{
            \node[shape=circle,draw,inner sep=0.3pt] (char) {#1};}}
            
\newcommand{\cmark}{\ding{51}}%
\newcommand{\xmark}{\ding{55}}%

\newcommand{\nosection}[1]{\vspace{2pt}\noindent\textbf{#1\ }}

\definecolor{commentcolor}{RGB}{110,154,155}   
\newcommand{\PyComment}[1]{\ttfamily\textcolor{commentcolor}{\# #1}}  
\newcommand{\PyCode}[1]{\ttfamily\textcolor{black}{#1}} 

\definecolor{linkcol}{RGB}{134,22,87} 
\definecolor{citecol}{RGB}{22,91,137} 
\definecolor{urlcol}{RGB}{20,88,21}
\hypersetup{
   colorlinks=true,
   linkcolor=linkcol,
   citecolor=citecol,
   urlcolor=urlcol,
}

\definecolor{red}{rgb}{0.863, 0.075, 0.235}
\definecolor{background}{RGB}{249, 250, 250}
\definecolor{string}{RGB}{153, 199, 180}
\definecolor{comment}{RGB}{153, 153, 153}
\definecolor{normal}{RGB}{50, 50, 50}
\definecolor{identifier}{RGB}{102, 153, 204}
\definecolor{number}{RGB}{249, 174, 87}
\definecolor{green}{RGB}{0, 128, 0}
\definecolor{lightgray}{rgb}{0.93,0.93,0.93}
\definecolor{pink}{rgb}{0.858, 0.188, 0.478}
\definecolor{cyberpink}{RGB}{235,110, 152}

\newcommand{\ours}[0]{\texttt{lrGAE}\xspace}
\newcommand{\code}[0]{\url{https://github.com/EdisonLeeeee/lrGAE}\xspace}

\newcolumntype{g}{>{\columncolor{lightgray}}c}
\newcolumntype{x}{>{\columncolor{gray!70}}c}

\newcommand{\red}[1]{\textcolor{red}{#1}}
\newcommand{\blue}[1]{\textcolor{blue!70}{#1}}

\newcommand{\orange}[1]{\textcolor{orange}{#1}}
\newcommand{\pink}[1]{\textcolor{pink}{#1}}

\newcommand{\mygreen}[1]{\textcolor[RGB]{54,174,124}{\textbf{#1}}}

\definecolor{red2}{rgb}{0.863, 0.075, 0.235}
\definecolor{blue2}{rgb}{0.047, 0.365, 0.647}

\newcommand{\best}[1]{\cellcolor{cyberpink!30}{\textbf{#1}}}
\newcommand{\second}[1]{\cellcolor{blue2!30}{\underline{#1}}}

\newcommand{\leftg}[1]{\eqnmarkbox[red]{}{#1}}
\newcommand{\rightg}[1]{\eqnmarkbox[blue]{}{#1}}

\theoremstyle{plain}
\newtheorem{theorem}{Theorem}[section]

\newtheorem{lemma}[theorem]{Lemma}

\theoremstyle{definition}

\theoremstyle{remark}

\AtBeginDocument{%
  }

\copyrightyear{2026}
\acmYear{2026}
\setcopyright{cc}
\setcctype{by}
\acmConference[KDD '26]{Proceedings of the 32nd ACM SIGKDD Conference on Knowledge Discovery and Data Mining V.2}{August 09--13, 2026}{Jeju Island, Republic of Korea}
\acmBooktitle{Proceedings of the 32nd ACM SIGKDD Conference on Knowledge Discovery and Data Mining V.2 (KDD '26), August 09--13, 2026, Jeju Island, Republic of Korea}
\acmDOI{10.1145/3770855.3817903}
\acmISBN{979-8-4007-2259-2/2026/08}
\begin{document}

\title{Revisiting Graph Autoencoders as Implicit Contrastive Learners}


\author{Jintang Li}
\email{edisonlee@xmu.edu.cn}
\authornote{Corresponding author.}
\affiliation{%
  \institution{Key Laboratory of Multimedia
Trusted Perception and Efficient Computing, Ministry of Education of China, Xiamen University}
      \city{Xiamen}
  \country{China}
}

\author{Ruofan Wu}
\email{wuruofan1989@gmail.com}
\affiliation{\institution{Coupang}
      \city{Shanghai}
    \country{China}}

\author{Yuchang Zhu}
\email{zhuych27@mail2.sysu.edu.cn}
\affiliation{\institution{Sun Yat-sen University}
      \city{Guangzhou}
    \country{China}}

\author{Huizhe Zhang}
\email{zhanghzh33@mail2.sysu.edu.cn}
\affiliation{\institution{Sun Yat-sen University}
      \city{Guangzhou}
    \country{China}}

\author{Zulun Zhu}
\email{ZULUN001@ntu.edu.sg}
\affiliation{%
    \institution{Nanyang Technological University}
      \city{Singapore}
    \country{Singapore}
    }

\author{Liang Chen}
\email{chenliang6@mail.sysu.edu.cn}
\affiliation{\institution{Sun Yat-sen University}
      \city{Guangzhou}
    \country{China}}

\renewcommand{\shortauthors}{Jintang Li et al.}

\begin{abstract}
Graph autoencoders (GAEs) and graph contrastive learning (GCL) are two major paradigms for self-supervised representation learning on graphs, yet they are often studied in isolation and treated as fundamentally different approaches. In this work, we revisit GAEs through the lens of contrastive learning and show that both structure-based and feature-based GAEs can be conceptualized as implicitly graph contrastive learners. This perspective reveals that many existing GAEs differ primarily in how contrastive views are constructed, rather than in their learning objectives or architectures. Building on this insight, we introduce a unified formulation that highlights contrastive view design as a central and previously less explored dimension in GAEs. In particular, we identify asymmetric contrastive views, arising from mismatches in subgraph views, as an important yet underexplored design axis in prior GAE research. We formalize this insight within a unified framework and conduct systematic experiments on representative graph learning tasks to examine its impact on performance and efficiency. Our results show that interpreting GAEs as implicit contrastive learners offers a clearer understanding of existing models and provides practical guidance for designing effective and scalable graph autoencoders.
\end{abstract}



\maketitle

\section{Introduction}

In recent years, self-supervised learning (SSL) has emerged as a powerful learning paradigm for graph representation learning, approaching and sometimes even surpassing the performance of supervised counterparts on many downstream tasks~\cite{infomax,infonce}. Compared with the supervised paradigm, self-supervised learning gets equal or even better performance with limited or no-labeled data, which saves substantial annotation time and plenty of resources.
Overall, SSL purely makes use of rich unlabeled data via well-designed pretext tasks that exploit the underlying structure and patterns in the data.
This is particularly useful in scenarios where label annotations are prohibitively costly to acquire.
Most recent approaches are shaped by the design of pretext tasks and architectural design, which has led to two lines of research: contrastive and non-contrastive learning~\cite{GarridoCBNL23,BalestrieroL22}.

\begin{figure}[t]
    \centering
    \includegraphics[width=\linewidth]{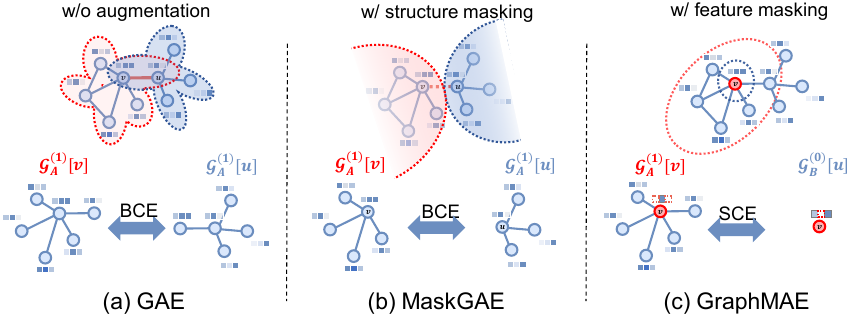}
    \caption{Technical comparison between GAE~\cite{gae}, MaskGAE~\cite{maskgae}, and GraphMAE~\cite{graphmae} from a contrastive learning perspective.}
    \label{fig:comparison}
    \vspace{-5mm}
\end{figure}

As one of the most successful and widespread SSL strategies, contrastive learning has first shown promising performance in vision representation learning~\cite{simclr,simcse}.
It brings together embeddings of different views of the same image while pushing away the embeddings from different ones. Contrastive learning develops rapidly and has recently been applied to the graph learning domain because of the scarcity of graph datasets with labels.
Contrastive learning on graphs (i.e., GCL)~\cite{graphcl} follows a similar framework to its counterparts in the vision domain, with the objective of maximizing the agreement between different graph augmentation views~\cite{wu2021self,spikegcl}. Basically, contrastive views are designed as nodes, subgraphs, or mixtures of the two that are either similar (positive pairs) or dissimilar (negative pairs)~\cite{dgi,suresh2021adversarial,graphcl,xu2021infogcl}.

On the other hand, non-contrastive learning approaches focus on capturing the generative aspects of graph data, with a promising line of research being graph autoencoders (GAEs)~\cite{gae}. Na\"ive GAEs adopt the edge-reconstruction principle to train the encoder, where edges of the input graph are expected to be reconstructed from hidden representations, thereby preserving the topological proximity and facilitating representation learning.
Compared to contrastive methods, self-supervised learning with GAEs is relatively less explored.
This is particularly due to the fact that GAEs have been criticized for their limitations in capturing complex graph structures and potentially overemphasizing proximity information~\cite{dgi,maskgae}.
As a result, the rise of graph contrastive learning has led to a shift away from traditional GAEs, as researchers seek more effective approaches for graph self-supervised learning. So far, contrastive methods have long become a dominant SSL paradigm on graphs.

Until recently, GAEs based on masked autoencoding have renewed interest in the field of graph SSL. Masked autoencoding~\cite{mae} is a technique where a subset of the components (e.g., graph topology or node features) in the graph is randomly masked or corrupted~\cite{graphmae,maskgae}. By learning to reconstruct the graph from the partially masked input, GAEs can better capture the underlying structure and semantic information encoded in the graph. Masked GAEs offer improved performance and have the potential to capture the underlying graph properties, making them a preferred choice in many advanced graph learning tasks.
Building on this momentum, research on (masked) GAEs has explored and become the new mainstream in graph self-supervised learning~\cite{gigamae,augmae,s2gae,graphmae2}.

While GAEs and contrastive approaches seem very different and have been described as such, we propose to take a closer look at the similarities between the two, both from a theoretical and empirical point of view. We argue that there exists a close relationship between GAEs and GCL. In this work, we revisit the GAEs studied in previous works and closely look into GAEs from a contrastive learning perspective. To be specific, we demonstrate that
\begin{tcolorbox}[width=1.0\linewidth, colframe=gray, colback=lightgray!30, boxsep=0mm, arc=2mm, left=2mm, right=2mm, top=2mm, bottom=2mm]
    {GAEs, whether employing structure or feature reconstruction, with or without masked autoencoding, implicitly perform \textit{graph contrastive learning} on two paired subgraph views.}
\end{tcolorbox}

The equivalence between structure-based GAEs and graph contrastive learning was initially demonstrated by~\cite{maskgae} and further extended in our work by additionally taking feature-based GAEs into consideration. To support our claim, we provide an illustrative comparison of three representative GAEs from a contrastive learning perspective in Figure~\ref{fig:comparison}. Essentially, vanilla GAE~\cite{gae} and MaskGAE~\cite{maskgae} are structure-based approaches that contrast a pair of connected nodes from their original and augmented structural views, respectively. On the other hand, GraphMAE~\cite{graphmae} is a feature-based approach that performs asymmetric graph contrastive learning by contrasting a node itself with its original and augmented subgraph views.

The above findings bridge the gap between graph contrastive and generative SSL methods, and further motivate us to revisit GAEs through a contrastive lens.
In particular, we propose \ours, which casts GAEs into an explicit contrastive architecture described by standard components, while centering the analysis on \emph{contrastive view construction} as the key design dimension that determines what information is aligned.
Under this unified view, we relate representative GAEs to \ours as special cases and highlight their connections through a common view-construction lens, which in turn motivates more effective contrastive-based GAE designs.

\paragraph{Contributions.}
This paper makes the following contributions.

\begin{itemize}
\item \textbf{A unifying perspective.}
We show that GAEs can be conceptualized as \emph{implicit contrastive learners}, where reconstruction objectives correspond to contrastive alignment between paired subgraph views. This perspective bridges reconstruction-based GAEs and modern GCL under a single formulation.

\item \textbf{An explicit design space.}
We introduce \ours, a modular framework that makes contrastive view construction in GAEs explicit along three orthogonal axes: graph view pairing, receptive-field asymmetry, and node pairing. This formulation recovers several existing GAEs as special cases and explores new \emph{asymmetric edge-paired} variants for GAEs.

\item \textbf{Empirical results.}
We instantiate three representative variants of \ours (\ours~\circled{6}/\circled{7}/\circled{8}) and evaluate them on multiple graph learning tasks. The results show that asymmetric designs achieve competitive performance with masked GAEs and strong GCL baselines, while offering favorable accuracy and efficiency trade-offs across diverse scenarios.
\end{itemize}

\section{Related Work}
To situate GAEs in a broader context, we discuss recent advances in graph self-supervised learning, including graph contrastive learning, graph autoencoders, and their recent masked variants.

\paragraph{Graph contrastive learning.}
GCL is a general self-supervised learning paradigm excelling at capturing invariant information from diverse graph augmentation views.
GCL has taken over the mainstream of self-supervised learning on graphs for years. Many works in this direction have recently flourished, with promising examples including DGI~\cite{dgi}, MVGRL~\cite{mvgrl}, and GRACE~\cite{grace}. While most GCLs potentially suffer from scalability issues due to complex augmentation and sampling strategies, several efforts have been made to scale up GCL through augmentation-free paradigms (e.g., AFGRL~\cite{afgrl}), architecture simplification (e.g., BGRL~\cite{bgrl} and SGCL~\cite{sgcl}) or in-batch feature decorrelation (e.g., CCA-SSG~\cite{cca_ssg}).

\paragraph{Graph autoencoders.}
GAEs are one such non-contrastive-based method that aims to learn meaningful representations by leveraging the graph reconstruction as the pretext task, i.e., by reconstructing certain inputs within a given context.
The pioneering works on GAEs can be traced back to GAE and VGAE~\cite{gae}, which utilize GNNs as the encoder and employ dot-product for link prediction decoding. Follow-up GAEs~\cite{seegera,argae,mgae,sigvae} mostly share a similar architecture, employing structure reconstruction or integrating both structure and feature reconstruction as their objectives. While GAEs typically excel in link-level tasks, they have been criticized for over-emphasizing proximity information at the expense of structural information and naturally underperform on node- and graph-level tasks pretrained on the graph reconstruction task~\cite{maskgae,graphmae}. This makes GAEs less preferable as a choice of graph self-supervised methods when compared to GCLs.

\paragraph{Masked graph autoencoders.}
Masked graph autoencoders are advanced GAEs that learn representations by reconstructing randomly masked patches from a graph input.
There are two major lines of masked GAEs, with each line primarily focusing on reconstruction at either the feature level~\cite{graphmae,graphmae2,gigamae,augmae} or the structure level~\cite{maskgae,s2gae}. Two seminal works include GraphMAE~\cite{graphmae} and MaskGAE~\cite{maskgae}.
GraphMAE introduces masked feature reconstruction on graphs as self-supervision, while MaskGAE focuses on graph structure reconstruction. The success of masked GAEs has spurred a surge of research in this area after the presentation.
Later works seek to improve masked GAEs through the utilization of powerful encoders~\cite{gmae}, advanced masking strategies~\cite{gigamae}, and regularization techniques~\cite{augmae,graphmae2}. However, a unified framework for GAEs and their masked alternatives is lacking, which motivates us to address this gap in our work.

\section{GAEs: generative yet contrastive}
\label{sec:gae_contrastive}

\subsection{Graph contrastive learning}
Let $\mathcal{G}=(\mathcal{V}, \mathcal{E})$ be an attributed graph with $\mathcal{V}=\{v_i\}^N$ and $\mathcal{E} \subseteq \mathcal{V}\times \mathcal{V}$ the set of nodes and edges, respectively. $N$ is the number of nodes in the graph.
GCL commonly involves generating two augmented views, denoted as $\mathcal{G}_A$ and $\mathcal{G}_B$, and trying to maximize the mutual information or the correspondence between two different views to train the encoder $f_\theta$. Here, $f_\theta$ represents an encoder network that maps the graph structure and node features simultaneously into a low-dimensional space. In this context, the goal of maximizing mutual information is achieved by minimizing the following objective:
\begin{equation}
    \label{eq:gcl}
    \min_{\theta} \frac{1}{|\mathcal{V}|} \sum_{v\in\mathcal{V}}\mathcal{L}\left({\mathbf{Z}_A[v]}, {\mathbf{Z}_B[v]}\right),
\end{equation}
where $\mathbf{Z}_A = f_\theta(\mathcal{G}_A), \mathbf{Z}_B=f_\theta(\mathcal{G}_B)$, and $\mathcal{L}$ refers to the contrastive loss, such as InfoNCE~\cite{infonce}. Typically, $f_\theta$ is a GNN network with receptive fields of $k$ (e.g., the depth of the network). We can further simplify Eq.~\ref{eq:gcl} as follows:
\begin{equation}
    \label{eq:gcl2}
    \min_{\theta} \frac{1}{|\mathcal{V}|} \sum_{v\in\mathcal{V}}\mathcal{L}\left({{\mathcal{G}^{(k)}_A[v]}}, {\mathcal{G}^{(k)}_B[v]}\right),
\end{equation}
where $\mathcal{G}^{(k)}_A[v]$ denotes the receptive fields of node $v$ in graph $\mathcal{G}$, with $f_\theta$ implicitly defined. In this way, we have a compact representation of GCL on its core component, i.e., the contrastive pair $\mathcal{G}^{(k)}_A[v]$ and $\mathcal{G}^{(k)}_B[v]$. In what follows, we will further generalize the representation of GAEs to adopt the form of this contrastive formulation.

\subsection{Graph autoencoders}
Technically, GAEs are encoding-decoding architectures that follow the graph-reconstruction principle as self-supervision. The goal of GAEs is to reconstruct or decode graph components, such as edges or features, from hidden representations. A typical GAE consists of an encoder network $f_\theta$, similar to GCLs, which learns low-dimensional representations, as well as a decoder network $g_\phi$ that performs graph reconstruction pretext tasks. Here we first introduce the learning objective of the conventional GAE~\cite{gae}, which is to reconstruct the graph structure, following the form described in~\cite{maskgae}:
\begin{align}
    \label{eq:gae1}
     & \mathcal{L}^+  =\frac{1}{|\mathcal{E}^+|}\sum_{(u, v)\in \mathcal{E}^+}\log g_\phi(\mathbf{Z}[u], \mathbf{Z}[v])                                     \\
    \label{eq:gae12}
     & \mathcal{L}^-  = \frac{1}{|\mathcal{E}^-|}\sum_{(u^\prime, v^\prime)\in \mathcal{E}^-}\log(1-g_\phi(\mathbf{Z}[{u^\prime}], \mathbf{Z}[{v^\prime}])) \\
     & \mathcal{L}_\text{GAE}    = - \left( \mathcal{L}^+ + \mathcal{L}^- \right),
\end{align}
where $\mathcal{E}^+$ is a set of positive edges and is usually a subset of $\mathcal{E}$, i.e., $\mathcal{E}^+\subseteq\mathcal{E}$. Correspondingly, $\mathcal{E}^-$ is a set of negative edges sampled from the graph and $\mathcal{E}^+\cap \mathcal{E}^-=\emptyset$. $\mathbf{Z}=f_\theta(\mathcal{G})$ in which $f_\theta$ is a GNN network such as GCN~\cite{kipf2016semi} or GAT~\cite{velickovic2018graph}. $g_\phi$ is the decoder network, which can be a simple dot-product or an advanced neural network performed on the combined representations:
\begin{equation}
    \label{eq:decode}
    g_\phi(x,y) = x \cdot y \ \ \text{or} \ \ g_\phi(x,y) = \text{MLP}_\phi(x\|y),
\end{equation}
where $\text{MLP}_\phi$ denotes a multi-layer perceptron (MLP) network parameterized by $\phi$; $\|$ is the concatenate operation.
Without loss of generality, we refer to GAEs that follow the form of Eq.~\ref{eq:gae1} as \textit{structure-based GAEs}.

Literature has shown that the structure-based GAEs might over-emphasize proximity information that is not always beneficial for self-supervised learning~\cite{dgi,maskgae,grace}. There are also works inheriting from VAE~\cite{vae} that seek to reconstruct the graph from its feature perspective:
\begin{equation}
    \label{eq:gae2}
    \mathcal{L} = \frac{1}{|\mathcal{P}|}\sum_{(i, j)\in \mathcal{P}} (g_\phi(\mathbf{Z})[i,j]-\mathbf{X}[i,j])^2,
\end{equation}
where $\mathcal{P}$ is the set of coordinates denoting the elements in the feature matrix to be reconstructed. $g_\phi$ is also a decoder network defined by an MLP on the input representations, i.e., $g_\phi(x) = \text{MLP}_\phi(x)$. Similarly, we term GAEs following the form of Eq.~\ref{eq:gae2} as \textit{feature-based GAEs} or GAE$_f$ for short.

\subsection{Connecting GAEs to GCLs}
\input{connection_to_gcl}

\section{\ours: design space for GAEs from a contrastive learning perspective}

We have shown that GAEs can be viewed as generative yet implicitly contrastive models. 
In this section, we introduce \ours (\underline{l}eft-\underline{r}ight contrastive \texttt{GAE}) as a unifying formulation to analyze and instantiate contrastive view designs in GAEs, rather than as a new modeling paradigm.
Following the works in GCLs~\cite{grace,mvgrl,graphcl}, we decompose the design space of \ours from five key dimensions: (1) augmentations, (2) contrastive views, (3) encoder/decoder networks, (4) contrastive loss, and optionally, (5) negative samples. Among these, the \textbf{contrastive views} distinguish GAEs from the line of GCL methods. 
In this section, we do not revisit these standard components; instead, we center our analysis on contrastive view construction, which is the dimension that most directly determines how GAEs instantiate implicit contrastive signals.

\begin{table*}[ht]
    \centering
    \small
    \vspace{-1mm}
    \caption{A taxonomy of contrastive view configurations in GAEs, expressed through the \ours formulation.}
    \label{tab:cases}
    \begin{threeparttable}{
            {
                    \begin{tabular}{lxccccggg}
                        \toprule
                        \multicolumn{9}{c}{\textbf{Contrastive views:} $\leftg{{\mathcal{G}^{(l)}_A[v]}}\leftrightarrow\rightg{{\mathcal{G}^{(r)}_B[u]}}$}                                                            \\
                        \cmidrule{2-9}
                                                  & \circled{1} & \circled{2} & \circled{3}        & \circled{4} & \circled{5} & \circled{6}                & \circled{7}                & \circled{8}                \\
                        \midrule
                        \textbf{Graph views}      & $A=B$       & $A\neq B$   & $A=B$              & $A\neq B$   & $A=B$       & $A=B$                      & $A\neq B$                  & $A\neq B$                  \\
                        \textbf{Receptive fields} & $l=r$       & $l=r$       & $l\neq r$          & $l\neq r$   & $l=r$       & $l \neq r$                 & $l=r$                      & $l\neq r$                  \\
                        \textbf{Node pairs}       & $v=u$       & $v=u$       & $v=u$              & $v=u$       & $v\neq u$   & $v \neq u$                 & $v\neq u$                  & $v\neq u$                  \\
                        \textbf{Abbreviation}     & AAllvv      & ABllvv      & AAlrvv             & ABlrvv      & AAllvu      & AAlrvu                     & ABllvu                     & ABlrvu                     \\
                        \textbf{Implementations}  & N/A         & GCLs~\cite{grace,graphcl}        & GAE$_f$ $^\dagger$~\cite{gae} & GraphMAE~\cite{graphmae}    & MaskGAE~\cite{maskgae}     & \textbf{?} & \textbf{?} & \textbf{?} \\
                        \bottomrule
                    \end{tabular}
                }
            \begin{tablenotes}\scriptsize
                \item[$\dagger$]We refer to the feature-based variant (see Eq.~\ref{eq:gae2}) of GAE here.
            \end{tablenotes}
        }
    \end{threeparttable}
    \vspace{-3mm}
\end{table*}

\subsection{Contrastive views in GAEs}
General GCL methods typically adopt a dual-branch architecture that maximizes agreement between two views of the same instance (e.g., a graph and its augmentation). In contrastive learning, the choice of views determines what information is preserved and therefore strongly influences the learned representations. As in Eq.~\ref{eq:gcl2}, a node-centric contrastive pair in GCL can be denoted as ${\mathcal{G}^{(l)}_A[v]} \leftrightarrow {\mathcal{G}^{(r)}_B[v]}$, which aligns the $k$-hop neighborhood representations of node $v$ across two graph views $\mathcal{G}_A$ and $\mathcal{G}_B$ (with $k$ typically induced by a $k$-layer GNN). This provides a standard way to contrast two views of a graph.

\paragraph{Unified objective.}
For GAEs, a similar spirit applies, except that the paired views can be naturally induced by graph structure and the reconstruction targets. We therefore extend the GCL view notation and formalize \ours as learning from \emph{paired subgraph views}:
\begin{equation}
\label{eq:gae_gcl}
\min_{\theta}\;
\frac{1}{|\mathcal{D}_{\mathcal{P}}|}
\sum_{(v,u)\in \mathcal{D}_{\mathcal{P}}}
\mathcal{L}\left(\leftg{{\mathcal{G}^{(l)}_A[v]}}, \rightg{\mathcal{G}^{(r)}_B[u]}\right),
\end{equation}
where $l, r \in {1, \dots, k}$ denote receptive fields constrained by the encoder depth $k$, and $\mathcal{G}_A$ and $\mathcal{G}_B$ are the paired graph views. $\mathcal{D}*{\mathcal{P}}$ denotes the set of node pairs induced by the pair set $\mathcal{P}$. When $\mathcal{P}$ samples identical-node pairs ($u = v$), the objective is node-centric; when $\mathcal{P}$ samples edge-endpoint pairs ($v \neq u$ with $(v,u) \in \mathcal{E}$), the objective becomes edge-centric. More generally, $\mathcal{L}$ can be implemented with contrastive objectives (e.g., InfoNCE) or reconstruction objectives used by GAEs; we defer the concrete implementations to Appendix~\ref{app:impl}.

In a nutshell, \ours makes explicit how paired views are formed through three orthogonal components: (i) \emph{graph views} ($\mathcal{G}_A$ versus $\mathcal{G}_B$), (ii) \emph{receptive fields} ($l$ versus $r$), and (iii) \emph{node pairs} ($u$ versus $v$). Enumerating the binary choices along these axes yields $2^3$ view cases, summarized in Table~\ref{tab:cases}, which also maps representative existing methods to the corresponding cases. This perspective clarifies how different GAE variants align with GCL principles and highlights under-explored configurations within the design space.
We treat Case~\circled{1} as degenerate because it aligns identical views and typically provides little learning signal under standard contrastive objectives. Since Cases~\circled{2}--\circled{5} have been widely explored in prior work, we focus on the under-studied asymmetric edge-paired cases (Cases~\circled{6}--\circled{8}) and investigate them within the \ours framework.

\subsection{\ours implementations and new variants}
\label{sec:new}

Most existing GAE-style methods instantiate GAEs in a symmetric or node-centric manner (e.g., Cases~\circled{2}--\circled{5}), leaving the asymmetric cases under-explored. In this work, we focus on Cases~\circled{6}, \circled{7}, and \circled{8}, which introduce asymmetry in receptive fields and/or graph views. We instantiate these cases within the \ours framework in Eq.~\ref{eq:gae_gcl}.

\paragraph{\ours~\protect\circled{6}: asymmetric receptive fields.}
\ours~\circled{6} extends the design of MaskGAE~\cite{maskgae} by allowing the receptive fields of the contrasted node representations to differ, i.e., $l \ne r$:
\begin{equation}
\label{eq:case6}
\ours~\circled{6}:\quad A=B,\;\; (v,u)\in\mathcal{E},\;\; v\neq u,\;\; l\neq r.
\end{equation}
Concretely, it aligns an $l$-hop view around one endpoint with an $r$-hop view around the other endpoint in the same graph view. This asymmetry allows the model to contrast local evidence against broader context across edge endpoints, which can mitigate overly local shortcuts and encourage multi-scale structural encoding.

\paragraph{\ours~\protect\circled{7}: asymmetric graph augmentations.}
\ours~\circled{7} is structurally similar to \ours~\circled{6}, but keeps the receptive fields identical while using different graph views. It instantiates an edge-paired view in the following manner:
\begin{equation}
\label{eq:case7}
\ours~\circled{7}:\quad A\neq B,\;\; (v,u)\in\mathcal{E},\;\; v\neq u,\;\; l=r.
\end{equation}
Compared with \ours~\circled{6}, this case shares the same edge pairing but introduces asymmetry through graph augmentations while keeping the receptive fields the same, which enables the two branches to encode complementary structural signals from differently perturbed graphs.

\paragraph{\ours~\protect\circled{8}: asymmetric receptive fields and graph views.}
\ours~\circled{8} is the most general asymmetric configuration among the three. It considers different graph views ($A \ne B$), unequal receptive fields ($l \ne r$), and different nodes connected by an edge ($u \ne v$ with $(v,u)\in\mathcal{E}$):
\begin{equation}
\label{eq:case8}
\ours~\circled{8}:\quad A\neq B,\;\; (v,u)\in\mathcal{E},\;\; v\neq u,\;\; l\neq r.
\end{equation}
It contrasts edge-endpoint representations across \emph{different} graph views and \emph{different} receptive fields, thereby combining view perturbation with multi-scale relational alignment. This design can provide richer contrastive signals for relational objectives such as link prediction, where aligning heterogeneous neighborhoods across endpoints may be more informative than symmetric, same-view pairings.

\begin{table*}[h]
    \centering
    \small
    \caption{Link prediction results (\%) on three citation networks.
        In each column, the best score is highlighted in pink and the second-best result is highlighted in blue.
    }
    {
        \begin{tabular}{llcccccccc}
            \toprule
                                                                             &                                       & \multicolumn{2}{c}{\textbf{Cora}}            &                                            & \multicolumn{2}{c}{\textbf{CiteSeer}} &                                            & \multicolumn{2}{c}{\textbf{PubMed}}                                                                                                                              \\
            \cmidrule{3-4}  \cmidrule{6-7}  \cmidrule{9-10}
                                                                             &                                       & AUC                                          & AP                                         &                                       & AUC                                        & AP                                           &                     & AUC                                          & AP                                           \\
            \midrule
            \multirow{7}{*}{\rotatebox{0}{\thead{\textbf{Standard GCL}}}}
                                                                             & DGI~\cite{dgi}                        & 86.0$_{\pm0.8}$                              & 86.4$_{\pm0.6}$                            &                                       & 89.5$_{\pm0.4}$                            & 89.1$_{\pm0.5}$                              &                     & 91.2$_{\pm0.6}$                              & 90.8$_{\pm0.7}$                              \\
                                                                             & GRACE~\cite{grace}                    & 87.1$_{\pm0.2}$                              & 85.9$_{\pm0.7}$                            &                                       & 90.4$_{\pm1.1}$                            & 90.0$_{\pm0.9}$                              &                     & 92.3$_{\pm0.4}$                              & 91.9$_{\pm0.6}$                              \\
                                                                             & BGRL~\cite{bgrl}                      & 89.5$_{\pm0.7}$                              & 89.0$_{\pm0.5}$                            &                                       & 85.8$_{\pm0.4}$                            & 85.3$_{\pm0.6}$                              &                     & 96.8$_{\pm0.1}$                              & 96.4$_{\pm0.2}$                              \\
                                                                             & CCA-SSG~\cite{cca_ssg}                & 92.1$_{\pm0.6}$                              & 92.8$_{\pm0.7}$                            &                                       & 92.5$_{\pm0.6}$                            & 92.1$_{\pm0.7}$                              &                     & 94.6$_{\pm0.3}$                              & 94.2$_{\pm0.4}$                              \\
                                                                             & GGD~\cite{ggd}                        & 91.4$_{\pm0.5}$                              & 92.0$_{\pm0.6}$                            &                                       & 93.5$_{\pm0.7}$                            & 93.1$_{\pm0.6}$                              &                     & 95.1$_{\pm0.4}$                              & 94.7$_{\pm0.5}$                              \\
                                                                             & S3GCL~\cite{s3gcl}                    & 92.7$_{\pm0.4}$                              & 91.7$_{\pm0.8}$                            &                                       & 94.2$_{\pm0.5}$                            & 93.8$_{\pm0.6}$                              &                     & 95.8$_{\pm0.3}$                              & 95.3$_{\pm0.4}$                              \\
                                                                             & EPAGCL~\cite{epagcl}                  & 89.9$_{\pm1.2}$                              & 88.6$_{\pm1.0}$                            &                                       & 92.1$_{\pm0.9}$                            & 91.6$_{\pm0.8}$                              &                     & 94.0$_{\pm0.6}$                              & 93.5$_{\pm0.7}$                              \\

            \midrule
            \multirow{7}{*}{\rotatebox{0}{\thead{\textbf{Feature-based}}}}   & GAE$_f$~\cite{gae}                    & 75.0$_{\pm1.2}$                              & 74.3$_{\pm0.9}$                            &                                       & 70.6$_{\pm2.4}$                            & 70.8$_{\pm2.1}$                              &                     & 82.2$_{\pm0.5}$                              & 81.3$_{\pm0.4}$                              \\
                                                                             & GraphMAE~\cite{graphmae}              & 93.9$_{\pm0.4}$                              & 94.1$_{\pm0.2}$                            &                                       & 93.8$_{\pm0.5}$                            & 94.9$_{\pm0.5}$                              &                     & 95.9$_{\pm0.3}$                              & 95.5$_{\pm0.2}$                              \\
                                                                             & GraphMAE2~\cite{graphmae2}            & 94.1$_{\pm0.5}$                              & 94.3$_{\pm0.4}$                            &                                       & 91.9$_{\pm0.5}$                            & 93.4$_{\pm0.6}$                              &                     & 95.6$_{\pm0.1}$                              & 95.2$_{\pm0.3}$                              \\
                                                                             & AUG-MAE~\cite{augmae}                 & 95.9$_{\pm0.6}$                              & 96.2$_{\pm0.3}$                            &                                       & 94.7$_{\pm0.2}$                            & 95.8$_{\pm0.3}$                              &                     & 94.5$_{\pm0.2}$                              & 94.0$_{\pm0.1}$                              \\
                                                                             & GiGaMAE~\cite{gigamae}                & 93.5$_{\pm0.5}$                              & 94.4$_{\pm0.7}$                            &                                       & 97.5$_{\pm0.4}$                            & \second{97.2$_{\pm0.1}$}                     &                     & 97.5$_{\pm0.3}$                              & 97.3$_{\pm0.4}$                              \\
                                                                             & SeeGera~\cite{seegera}                & 95.5$_{\pm0.3}$                              & 95.2$_{\pm0.6}$                            &                                       & 97.0$_{\pm0.6}$                            & {97.1$_{\pm0.4}$}                            &                     & 97.6$_{\pm0.5}$                              & 97.8$_{\pm0.6}$                              \\
            \midrule
            \multirow{6}{*}{\rotatebox{0}{\thead{\textbf{Structure-based}}}} & GAE~\cite{gae}                        & 92.5$_{\pm0.8}$                              & 93.7$_{\pm0.6}$                            &                                       & 88.4$_{\pm1.3}$                            & 90.5$_{\pm1.2}$                              &                     & 98.0$_{\pm0.2}$                              & 98.1$_{\pm0.3}$                              \\
                                                                             & S2GAE~\cite{s2gae}                    & 94.5$_{\pm0.5}$                              & 93.8$_{\pm0.4}$                            &                                       & 94.0$_{\pm0.2}$                            & 95.3$_{\pm0.3}$                              &                     & 98.3$_{\pm0.1}$                              & 98.2$_{\pm0.4}$                              \\
                                                                             & MaskGAE~\cite{maskgae}                & \best{96.8$_{\pm0.2}$}                       & \second{97.0$_{\pm0.3}$}                   &                                       & \second{97.6$_{\pm0.1}$}                   & \best{97.9$_{\pm0.1}$}                       &                     & 98.7$_{\pm0.1}$                              & \best{98.8$_{\pm0.0}$}                       \\

                                                                             & \cellcolor{gray!10} \ours~\circled{6} & \cellcolor{gray!10} 96.3$_{\pm0.6}$          & \cellcolor{gray!10} 96.1$_{\pm0.7}$        & \cellcolor{gray!10}                   & \cellcolor{gray!10} 96.5$_{\pm0.2}$        & \cellcolor{gray!10} 96.4$_{\pm0.2}$          & \cellcolor{gray!10} & \cellcolor{gray!10} 98.1$_{\pm0.1}$          & \cellcolor{gray!10} 97.7$_{\pm0.3}$          \\
                                                                             & \cellcolor{gray!10} \ours~\circled{7} & \cellcolor{gray!10} \second{96.4$_{\pm0.8}$} & \cellcolor{gray!10} \best{97.2$_{\pm0.4}$} & \cellcolor{gray!10}                   & \cellcolor{gray!10} \best{97.7$_{\pm0.3}$} & \cellcolor{gray!10} \second{97.2$_{\pm0.4}$} & \cellcolor{gray!10} & \cellcolor{gray!10} \best{98.9$_{\pm0.3}$}   & \cellcolor{gray!10} \best{98.8$_{\pm0.1}$}   \\
                                                                             & \cellcolor{gray!10} \ours~\circled{8} & \cellcolor{gray!10} 96.3$_{\pm0.5}$          & \cellcolor{gray!10} 96.2$_{\pm0.5}$        & \cellcolor{gray!10}                   & \cellcolor{gray!10} 97.1$_{\pm0.1}$        & \cellcolor{gray!10} \second{97.2$_{\pm0.5}$} & \cellcolor{gray!10} & \cellcolor{gray!10} \second{98.8$_{\pm0.1}$} & \cellcolor{gray!10} \second{98.4$_{\pm0.2}$} \\
            \bottomrule
        \end{tabular}
    }
    \label{tab:link_predict}
\end{table*}

\begin{table*}[t]
    \centering
    \small
    \caption{Node classification accuracy (\%) on seven benchmark datasets.
        \texttt{OOM}: out-of-memory on an NVIDIA 3090 Ti GPU with 24GB memory.
    }
    \label{tab:node_clas}
    \begin{threeparttable}{
            {
                    \begin{tabular}{llccccccc}
                        \toprule
                                                                                         &                                      & \textbf{Cora}                             & \textbf{CiteSeer}                           & \textbf{PubMed}                             & \textbf{Photo}                              & \textbf{Computer}                        & \textbf{CS}                                 & \textbf{Physics}                            \\
                        \midrule
                        \multirow{7}{*}{\rotatebox{0}{\thead{\textbf{Standard GCL}}}}
                                                                                         & DGI~\cite{dgi}                       & 82.3$_{\pm1.6}$                           & 70.8$_{\pm1.7}$                             & 79.8$_{\pm1.6}$                             & 91.6$_{\pm0.2}$                             & 83.9$_{\pm0.4}$                           & 92.1$_{\pm0.4}$                             & 94.8$_{\pm0.2}$                             \\
                                                                                         & GRACE~\cite{grace}                   & 82.9$_{\pm1.4}$                           & 72.1$_{\pm1.5}$                             & 81.6$_{\pm1.4}$                             & 92.1$_{\pm0.2}$                             & 86.2$_{\pm0.2}$                           & 92.5$_{\pm0.2}$                             & 95.0$_{\pm0.1}$                             \\
                                                                                         & BGRL~\cite{bgrl}                     & 82.8$_{\pm0.9}$                           & 72.4$_{\pm0.9}$                             & 82.0$_{\pm0.8}$                             & 93.2$_{\pm0.3}$                             & 88.3$_{\pm0.1}$                           & 92.4$_{\pm0.6}$                             & 94.7$_{\pm0.4}$                             \\
                                                                                         & CCA-SSG~\cite{cca_ssg}               & 83.5$_{\pm0.7}$                           & \second{73.1$_{\pm0.5}$}                      & 80.8$_{\pm0.3}$                             & 93.1$_{\pm0.1}$                             & 88.7$_{\pm0.2}$                           & 92.7$_{\pm0.8}$                             & 95.2$_{\pm0.5}$                             \\
                                                                                         & GGD~\cite{ggd}                       & 83.2$_{\pm0.6}$                           & 72.2$_{\pm1.5}$                             & 81.8$_{\pm0.8}$                             & 92.4$_{\pm0.3}$                             & 86.9$_{\pm0.3}$                           & 92.4$_{\pm0.1}$                             & 95.0$_{\pm0.4}$                             \\
                                                                                         & S3GCL~\cite{s3gcl}                   & 83.7$_{\pm1.2}$                           & 71.5$_{\pm0.9}$                             & 81.2$_{\pm0.4}$                             & \second{93.4$_{\pm0.2}$}                   & 89.6$_{\pm0.2}$                           & 92.5$_{\pm0.3}$                             & 95.3$_{\pm0.3}$                             \\
                                                                                         & EPAGCL~\cite{epagcl}                 & 83.6$_{\pm1.4}$                           & 70.9$_{\pm0.8}$                             & 82.0$_{\pm0.2}$                             & 92.0$_{\pm0.3}$                             & 86.4$_{\pm0.4}$                           & 92.6$_{\pm0.3}$                             & 95.2$_{\pm0.2}$                             \\
                        \midrule
                        \multirow{7}{*}{\rotatebox{0}{\thead{\textbf{Feature-based}}}}   & GAE$_f$~\cite{gae}                   & 57.7$_{\pm1.3}$                           & 47.5$_{\pm1.3}$                             & 60.0$_{\pm0.6}$                             & 80.1$_{\pm0.2}$                             & 70.9$_{\pm0.3}$                           & 78.0$_{\pm0.4}$                             & 55.6$_{\pm0.7}$                             \\
                                                                                         & GraphMAE~\cite{graphmae}             & \best{84.5$_{\pm1.1}$}                    & 72.5$_{\pm0.9}$                             & 81.0$_{\pm0.8}$                             & 93.3$_{\pm0.2}$                             & \best{89.8$_{\pm0.2}$}                    & 92.7$_{\pm0.4}$                             & 95.1$_{\pm0.2}$                             \\
                                                                                         & GraphMAE2~\cite{graphmae2}           & 83.7$_{\pm2.1}$                           & 71.7$_{\pm1.5}$                             & 80.4$_{\pm1.0}$                             & 93.2$_{\pm0.5}$                             & 89.6$_{\pm0.4}$                           & 92.7$_{\pm0.5}$                             & 94.9$_{\pm0.1}$                             \\
                                                                                         & AUG-MAE~\cite{augmae}                & 83.9$_{\pm1.4}$                           & \second{73.1$_{\pm2.1}$}                      & 80.3$_{\pm1.5}$                             & \second{93.4$_{\pm0.8}$}                    & 89.6$_{\pm0.5}$                           & \texttt{OOM}                                & \texttt{OOM}                                \\
                                                                                         & GiGaMAE~\cite{gigamae}               & 83.2$_{\pm1.5}$                           & 69.8$_{\pm1.8}$                             & \second{82.5$_{\pm1.5}$}                    & \best{93.5$_{\pm0.5}$}                      & \second{89.7$_{\pm0.7}$}                  & 92.4$_{\pm0.7}$                             & \texttt{OOM}                                \\
                                                                                         & SeeGera~\cite{seegera}               & 84.0$_{\pm1.6}$                           & 72.6$_{\pm1.4}$                             & {80.4$_{\pm1.0}$}                           & {92.8$_{\pm0.5}$}                           & {88.3$_{\pm0.3}$}                         & 92.8$_{\pm0.2}$                             & 95.2$_{\pm0.1}$                             \\
                        \midrule
                        \multirow{6}{*}{\rotatebox{0}{\thead{\textbf{Structure-based}}}} & GAE~\cite{gae}                       & 79.8$_{\pm2.5}$                           & 64.9$_{\pm2.7}$                             & 76.2$_{\pm1.9}$                             & 89.9$_{\pm0.5}$                             & 79.2$_{\pm0.3}$                           & 92.0$_{\pm0.5}$                             & 95.2$_{\pm0.1}$                             \\
                                                                                         & S2GAE~\cite{s2gae}                   & 84.1$_{\pm1.4}$                           & 72.3$_{\pm1.5}$                             & 81.5$_{\pm0.8}$                             & 93.1$_{\pm0.3}$                             & 89.1$_{\pm0.4}$                           & 92.6$_{\pm0.1}$                             & 95.5$_{\pm0.2}$                             \\
                                                                                         & MaskGAE~\cite{maskgae}               & \second{84.4$_{\pm1.5}$}                  & 72.8$_{\pm0.7}$                             & \best{82.9$_{\pm0.4}$}                      & 93.0$_{\pm0.1}$                             & 89.5$_{\pm0.1}$                           & \second{92.9$_{\pm0.1}$}                    & 95.6$_{\pm0.2}$                             \\
                                                                                         & \cellcolor{gray!10}\ours~\circled{6} & \cellcolor{gray!10}84.0$_{\pm1.5}$        & \cellcolor{gray!10}{73.0$_{\pm1.8}$} & \cellcolor{gray!10}81.9$_{\pm0.5}$          & \cellcolor{gray!10}\best{93.5$_{\pm0.3}$}   & \cellcolor{gray!10}89.1$_{\pm0.3}$        & \cellcolor{gray!10}\second{92.9$_{\pm0.4}$} & \cellcolor{gray!10}95.4$_{\pm0.1}$          \\
                                                                                         & \cellcolor{gray!10}\ours~\circled{7} & \cellcolor{gray!10}84.2$_{\pm1.8}$        & \cellcolor{gray!10}72.5$_{\pm1.1}$          & \cellcolor{gray!10}81.1$_{\pm0.8}$          & \cellcolor{gray!10}\second{93.4$_{\pm0.5}$} & \cellcolor{gray!10}\best{89.8$_{\pm0.2}$} & \cellcolor{gray!10}92.8$_{\pm0.2}$          & \cellcolor{gray!10}\second{95.7$_{\pm0.2}$} \\
                                                                                         & \cellcolor{gray!10}\ours~\circled{8} & \cellcolor{gray!10}\best{84.5$_{\pm1.4}$} & \cellcolor{gray!10}\best{73.4$_{\pm1.4}$}          & \cellcolor{gray!10}\second{82.5$_{\pm1.0}$} & \cellcolor{gray!10}93.2$_{\pm0.4}$          & \cellcolor{gray!10}\second{89.7$_{\pm0.3}$}        & \cellcolor{gray!10}\best{93.1$_{\pm0.2}$}   & \cellcolor{gray!10}\best{95.8$_{\pm0.1}$}   \\
                        \bottomrule
                    \end{tabular}
                }
        }
    \end{threeparttable}
\end{table*}

\begin{table*}[t]
    \centering
    \small
    \caption{Link Prediction and node classification results (\%) on large-scale graphs.}  \label{tab:large}
    \begin{tabular}{llccc|ccc}
        \toprule
                                                                         &                                      & \multicolumn{3}{c}{\textbf{Link Prediction}} & \multicolumn{3}{c}{\textbf{Node Classification}}                                                                                                                                                                            \\
        \cmidrule{2-8}
                                                                         &                                      & \textbf{ogbl-ddi}                            & \textbf{ogbl-collab}                             & \textbf{ogbl-ppa}                         & \textbf{ogbn-arXiv}                & \textbf{ogbn-MAG}                         & \textbf{ogbn-Products}                      \\
                                                                         &                                      & Hits@20                                      & Hits@50                                          & Hits@10                                   & Accuracy                           & Accuracy                                  & Accuracy                                    \\
        \midrule
        \multirow{7}{*}{\rotatebox{0}{\thead{\textbf{Standard GCL}}}}
                                                                         & DGI~\cite{dgi}                       & \texttt{N/A}                                 & \texttt{OOM}                                     & \texttt{N/A}                              & 65.1$_{\pm0.4}$                    & 31.4$_{\pm0.3}$                           & \texttt{OOM}                                \\
                                                                         & GRACE~\cite{grace}                   & \texttt{N/A}                                 & \texttt{OOM}                                     & \texttt{N/A}                              & 68.7$_{\pm0.4}$                    & 31.5$_{\pm0.3}$                           & 79.5$_{\pm0.6}$                             \\
                                                                         & BGRL~\cite{bgrl}                     & \texttt{N/A}                                 & 21.6$_{\pm1.9}$                                  & \texttt{N/A}                              & 71.6$_{\pm0.1}$                    & 31.1$_{\pm0.1}$                           & 78.6$_{\pm0.0}$                             \\
                                                                         & CCA-SSG~\cite{cca_ssg}               & \texttt{N/A}                                 & 23.4$_{\pm1.2}$                                  & \texttt{N/A}                              & 71.2$_{\pm0.2}$                    & 31.8$_{\pm0.4}$                           & 75.3$_{\pm0.1}$                             \\
                                                                         & GGD~\cite{ggd}                       & \texttt{N/A}                                 & 25.6$_{\pm0.5}$                                  & \texttt{N/A}                              & 71.6$_{\pm0.5}$                    & 31.7$_{\pm0.7}$                           & 75.7$_{\pm0.4}$                             \\
                                                                         & S3GCL~\cite{s3gcl}                   & \texttt{N/A}                                 & 24.5$_{\pm1.4}$                                  & \texttt{N/A}                              & 71.3$_{\pm0.6}$                    & 30.9$_{\pm0.5}$                           & 81.4$_{\pm0.2}$                             \\
                                                                         & EPAGCL~\cite{epagcl}                 & \texttt{N/A}                                 & \texttt{OOM}                                     & \texttt{N/A}                              & 69.2$_{\pm0.1}$                    & \texttt{OOM}                              & \texttt{OOM}                                \\
        \midrule
        \multirow{6}{*}{\rotatebox{0}{\thead{\textbf{Feature-based}}}}   & GAE$_f$~\cite{gae}                              & \texttt{N/A}                                 & 20.2$_{\pm1.0}$                                  & \texttt{N/A}                              & 64.2$_{\pm0.7}$                    & 25.7$_{\pm0.4}$                           & 73.5$_{\pm0.6}$                             \\
                                                                         & GraphMAE~\cite{graphmae}                             & \texttt{N/A}                                 & 24.8$_{\pm1.1}$                                  & \texttt{N/A}                              & 71.0$_{\pm0.4}$                    & 32.2$_{\pm0.3}$                           & 78.9$_{\pm0.4}$                             \\
                                                                         & GraphMAE2~\cite{graphmae2}                            & \texttt{N/A}                                 & 23.2$_{\pm1.4}$                                  & \texttt{N/A}                              & \second{71.6$_{\pm0.2}$}           & 32.7$_{\pm0.2}$                           & 81.0$_{\pm0.2}$                             \\
                                                                         & AUG-MAE~\cite{augmae}                              & \texttt{N/A}                                 & \texttt{OOM}                                     & \texttt{N/A}                              & \texttt{OOM}                       & \texttt{OOM}                              & \texttt{OOM}                                \\
                                                                         & GiGaMAE~\cite{gigamae}                              & \texttt{N/A}                                 & \texttt{OOM}                                     & \texttt{N/A}                              & \texttt{OOM}                       & \texttt{OOM}                              & \texttt{OOM}                                \\
                                                                         & SeeGera~\cite{seegera}                              & \texttt{N/A}                                 & \texttt{OOM}                                     & \texttt{N/A}                              & \texttt{OOM}                       & \texttt{OOM}                              & \texttt{OOM}                                \\
        \midrule
        \multirow{6}{*}{\rotatebox{0}{\thead{\textbf{Structure-based}}}} & GAE~\cite{gae}                                  & 37.1$_{\pm3.2}$                              & 45.8$_{\pm1.7}$                                  & 2.3$_{\pm0.3}$                            & 68.9$_{\pm0.3}$                    & 30.1$_{\pm0.3}$                           & 77.8$_{\pm0.3}$                             \\
                                                                         & S2GAE~\cite{s2gae}                                & \second{65.8$_{\pm7.4}$}                     & 51.3$_{\pm1.2}$                                  & \second{3.4$_{\pm0.9}$}                   & 70.5$_{\pm0.2}$                    & 31.7$_{\pm0.1}$                           & 80.0$_{\pm0.5}$                             \\
                                                                         & MaskGAE~\cite{maskgae}                              & 64.7$_{\pm4.8}$                              & 53.2$_{\pm2.1}$                                  & 3.0$_{\pm0.8}$                            & 71.2$_{\pm0.1}$                    & \second{32.8$_{\pm0.2}$}                  & 79.6$_{\pm0.4}$                             \\
                                                                         & \cellcolor{gray!10}\ours~\circled{6} & \cellcolor{gray!10}64.9$_{\pm5.1}$           & \cellcolor{gray!10}52.7$_{\pm1.8}$               & \cellcolor{gray!10}2.8$_{\pm0.3}$         & \cellcolor{gray!10}70.8$_{\pm0.2}$ & \cellcolor{gray!10}32.7$_{\pm0.1}$        & \cellcolor{gray!10}80.7$_{\pm0.1}$          \\
                                                                         & \cellcolor{gray!10}\ours~\circled{7} & \cellcolor{gray!10}64.2$_{\pm6.3}$           & \second{53.5$_{\pm1.5}$}                         & \cellcolor{gray!10}3.1$_{\pm0.5}$         & \cellcolor{gray!10}71.4$_{\pm0.1}$ & \cellcolor{gray!10}\best{32.9$_{\pm0.1}$} & \cellcolor{gray!10}\second{81.9$_{\pm0.3}$} \\
                                                                         & \cellcolor{gray!10}
        \ours~\circled{8}                                                & \best{66.1$_{\pm5.4}$}               & \best{54.4$_{\pm1.2}$}                       & \best{3.8$_{\pm0.2}$ }                           & \cellcolor{gray!10}\best{71.9$_{\pm0.1}$} & \cellcolor{gray!10}32.2$_{\pm0.1}$ & \cellcolor{gray!10}\best{82.3$_{\pm0.2}$}                                               \\
        \bottomrule
    \end{tabular}
\end{table*}

\section{Experiments}
In this section, we conduct extensive experiments on multiple graph learning tasks to evaluate the performance of \ours. Specifically, our main experiments are conducted on ten graph datasets, including Cora, CiteSeer, PubMed~\cite{sen2008collective}, Photo, Computer~\cite{shchur2018pitfalls}, CS, Physics~\cite{shchur2018pitfalls}, and three large-scale graphs from the Open Graph Benchmark~\cite{hu2020ogb}. Our baselines span two lines of research: graph contrastive learning methods and graph autoencoders. 
For each baseline, we follow official implementations when available and keep encoder capacity comparable; all methods share the same splits and downstream protocol.
Due to space limitations, we refer readers to Appendix~\ref{appendix:setting} for details on the datasets, baseline methods, evaluation protocols, and the implementation of \ours. The source code for \ours, along with the baselines and scripts necessary to reproduce our experiments, is made available at \code.

\subsection{Main results}

\nosection{Link prediction.}
Table~\ref{tab:link_predict} presents the link prediction results in three citation graphs. We follow exactly the experimental settings in \cite{gae} and report the averaged AUC and AP across 10 runs. For traditional GCL and feature-based GAEs that are not trained with a structural decoder, we use dot-product operation over the learned node representations to perform link prediction. As can be observed, masked GAEs are more powerful than vanilla GAEs. The results are in line with the conclusions of previous works~\cite{maskgae,graphmae,augmae,s2gae} that augmentations such as masking can greatly benefit self-supervised learning.
In addition, structure-based GAEs, i.e., GAE, S2GAE, MaskGAE, and \ours~\circled{6}/\circled{7}/\circled{8}, generally achieve better performance on all datasets. By contrast, feature-based GAEs and GCL methods underperform in such a link prediction task due to the gap between the pretext task (e.g., regression) and the downstream task (classification) of feature-based GAEs. This highlights the importance of the generality of pretext tasks during self-supervised learning. It is worth noting that although GiGaMAE adopts a feature-reconstruction objective, it also achieves comparable performance with structure-based GAEs. The possible reason might be that the contrastive learning loss offers better generalization than regression loss (e.g., MSE and SCE~\cite{graphmae}).

\nosection{Node classification.}
To provide a comprehensive evaluation, we conduct node classification on seven widely used graph datasets.
For Cora, CiteSeer, and PubMed, we adopt the public splits following prior work~\cite{kipf2016semi, maskgae}. For the remaining datasets, we follow the recommended 8:1:1 train/validation/test random splits~\cite{shchur2018pitfalls}.
Table~\ref{tab:node_clas} reports the results over ten runs.
We first compare \ours against standard GCL baselines. Overall, modern GCL methods provide strong performance, especially on citation networks, confirming the effectiveness of agreement maximization between augmented views. However, their gains are not consistent across datasets, suggesting that contrastive view choices alone may be insufficient to fully exploit the structural supervision.
We then examine GAE-style methods. It is observed that masked GAEs yield equally good performance, particularly in Photo, Computer, and CS. However, most feature-based GAEs suffer from scalability issues in Physics, which has a large feature dimensionality. This reveals the potential limitation of feature-based GAEs, which are less flexible and scalable compared to structure-based GAEs.
We can also see that \ours~\circled{6}/\circled{7}/\circled{8}are consistently competitive with strong masked-GAE baselines and often achieve the best accuracy–efficiency trade-off among structure-based methods. \ours unleashes the power of GAEs with GCL principles and provides deep insights to facilitate self-supervised learning.

\begin{figure}[t]
    \centering
    \begin{minipage}{\linewidth}
        \centering
        \includegraphics[width=0.7\linewidth]{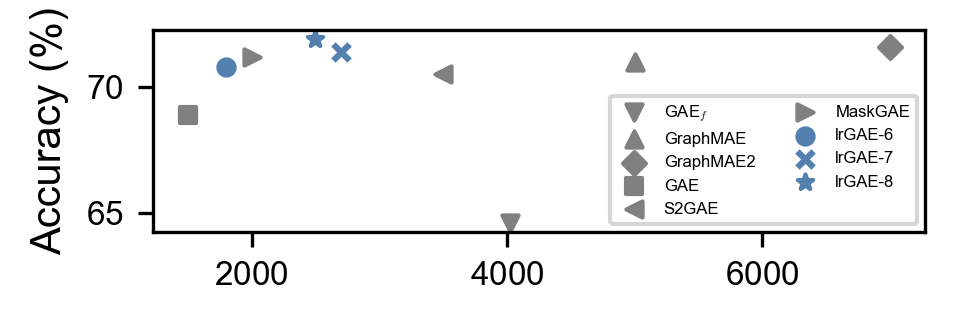}
    \end{minipage}

    \vspace{-3mm} 

    \begin{minipage}{\linewidth}
        \centering
        \includegraphics[width=0.7\linewidth]{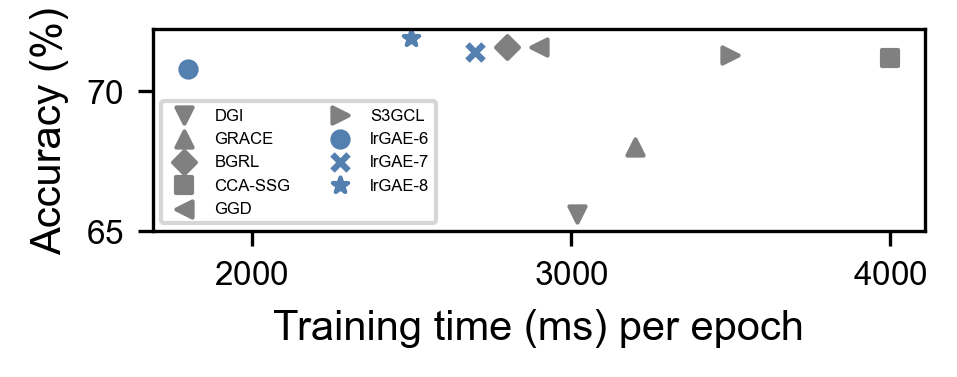}
    \end{minipage}

    \caption{Speed (ms) versus accuracy of different methods on the ogbn-arXiv dataset.}
    \label{fig:scatter}
    \vspace{-4mm}
\end{figure}

\nosection{Scalability to large-scale graphs.}
We evaluate scalability on six large-scale OGB datasets for both link prediction and node classification in Table~\ref{tab:large}, where \texttt{N/A} indicates the method is not directly applicable under our protocol.
Overall, structure-based methods are more practical at large-scale settings, as they avoid reconstructing high-dimensional node features and thus remain stable in memory and training cost. This is consistent with the \texttt{OOM} cases of several feature-based masked reconstruction methods, although GraphMAE/GraphMAE2 can still run and achieve strong node classification accuracy.
Standard GCL baselines are competitive on node classification, but their scalability is less consistent across datasets and they are not directly applicable to the link prediction setting. This reveals a mismatch between node-centric contrastive objectives and edge-centric tasks.
In contrast, our asymmetric variants \ours~\circled{6}/\circled{7}/\circled{8} are consistently applicable and memory-efficient on both tasks.
Figure~\ref{fig:scatter} further shows that \ours achieves a favorable accuracy--speed trade-off: while GraphMAE/GraphMAE2 can be slightly better in accuracy, they incur higher training cost, whereas \ours attains competitive performance with substantially lower training time.

\subsection{Ablation studies}
\label{sec:abla}
In this section, we study how the key design factors in our \ours\ framework affect performance, with the goal of providing actionable guidance for instantiating \ours\ in practice.
Concretely, we ablate four orthogonal components in the \ours\ design space: (i) graph augmentations, (ii) encoder architectures, (iii) contrastive objectives, and (iv) negative sampling strategies.
We conduct node classification experiments on \textsc{Cora}, and report results averaged over 10 runs.
Due to space limitations, additional results and extended discussions are deferred to Appendix~\ref{appendix:exp}.

\begin{figure}[t]
    \subfigure[Feature-based]{\includegraphics[width=0.48\linewidth]{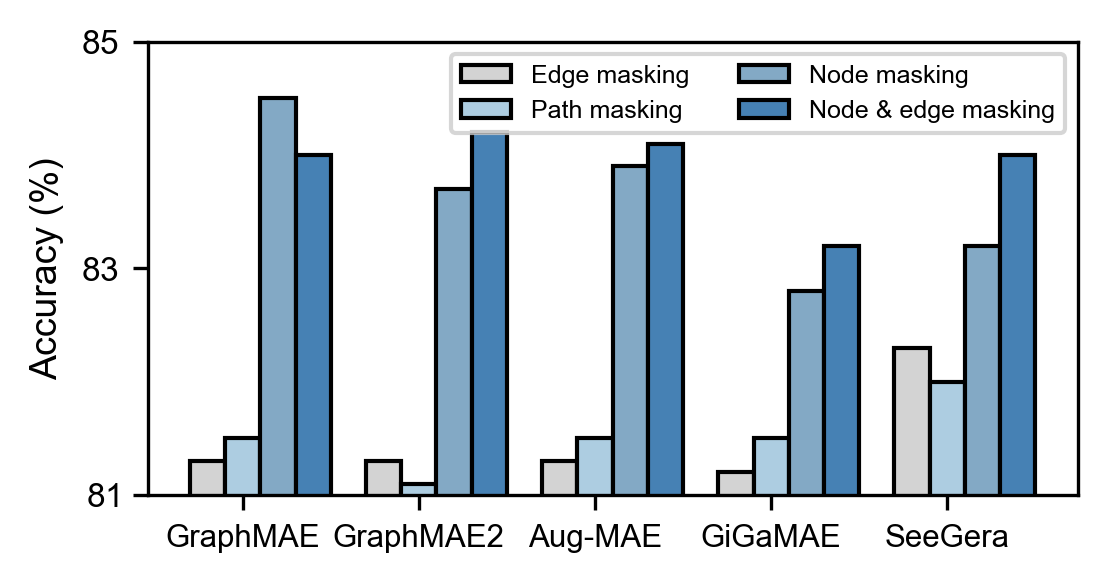}}
    \subfigure[Structure-based]{\includegraphics[width=0.48\linewidth]{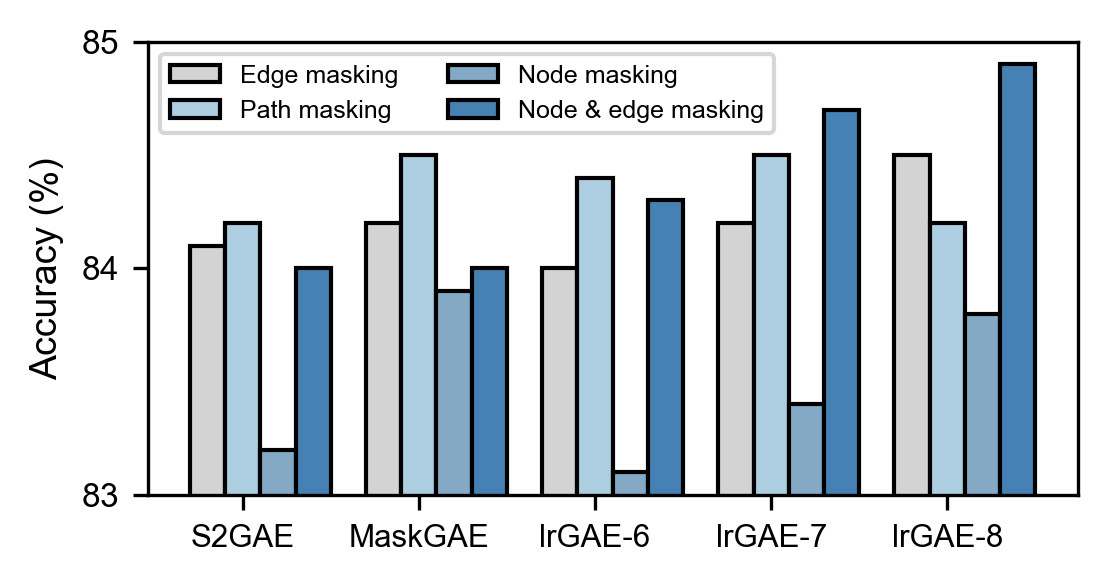}}
    \caption{Ablation on the augmentation strategies.}
    \label{fig:abla_aug}
\end{figure}

\nosection{Augmentations.}
We first conduct ablation studies on the augmentation techniques, which include edge masking~\cite{dropedge}, path masking~\cite{maskgae}, node masking~\cite{graphcl}, and the combination of node and edge masking. We set the masking ratio to 0.7 based on insights from the literature~\cite{graphmae,maskgae}. Figure~\ref{fig:abla_aug} presents the ablation results of structure-based and feature-based GAEs using various masking strategies. As observed, the performance of GAEs varies significantly with different masking strategies. In general, edge masking and path masking are better choices than node masking for structure-based GAEs, while node masking is more suitable for feature-based GAEs. This observation aligns with our intuition. In addition, node and edge masking together achieve the best performance in most cases, highlighting the effectiveness of combining masking techniques to enhance model performance.

\begin{figure}[t]
    \subfigure[Feature-based]{\includegraphics[width=0.48\linewidth]{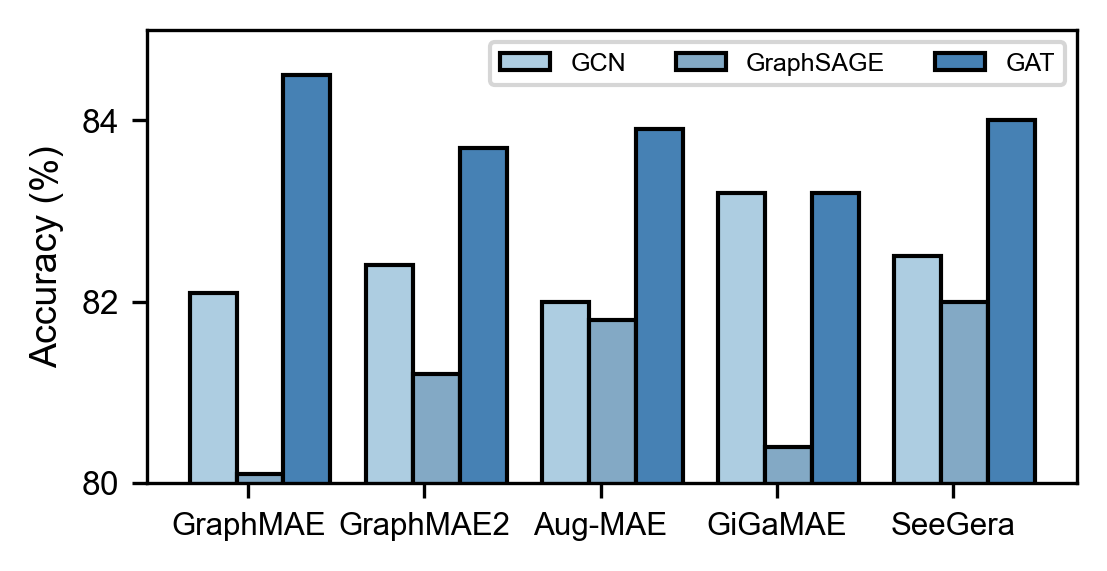}}
    \subfigure[Structure-based]{\includegraphics[width=0.48\linewidth]{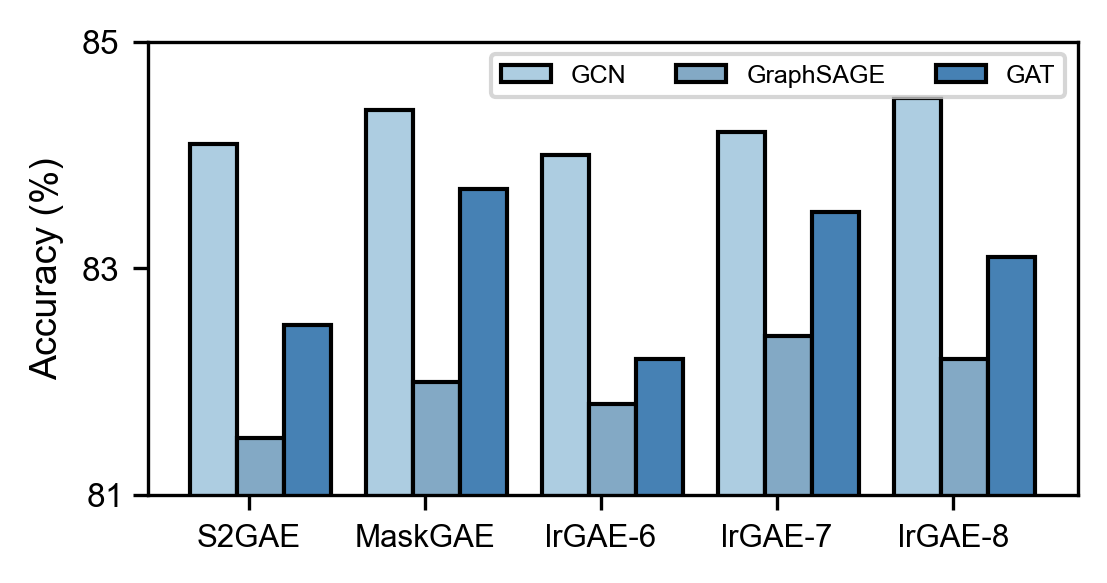}}
    \caption{Ablation on the encoder architectures.}
    \label{fig:abla_encoder}
\end{figure}

\nosection{Encoder networks.}
Given that most of the decoder networks in GAEs are MLPs, we focus the ablation experiments solely on the encoder networks. The encoder plays a crucial role in mapping graphs into low-dimensional representations. To investigate the potential of designing effective GAEs, we perform ablation studies using three different GNN encoders, including GCN~\cite{kipf2016semi}, GraphSAGE~\cite{hamilton2017inductive}, and GAT~\cite{velickovic2018graph}. The results, shown in Figure~\ref{fig:abla_encoder}, demonstrate that GCN and GAT are the most effective encoder architectures for structure-based and feature-based GAEs, respectively. Specifically, GCN excels at learning from the graph’s global structure, making it well-suited for structure-based GAEs, while GAT’s attention mechanism allows it to focus on the most relevant node features, making it effective for feature-based GAEs. In contrast, GraphSAGE shows slightly reduced performance across both structures, possibly due to its simpler aggregation function, which may not fully capture the complex relationships within the graph data. This observation is consistent with the conclusions of previous studies~\cite{maskgae,dgi,bgrl,gca,cca_ssg}.

\begin{figure}[t]
    \subfigure[Feature-based]{\includegraphics[width=0.48\linewidth]{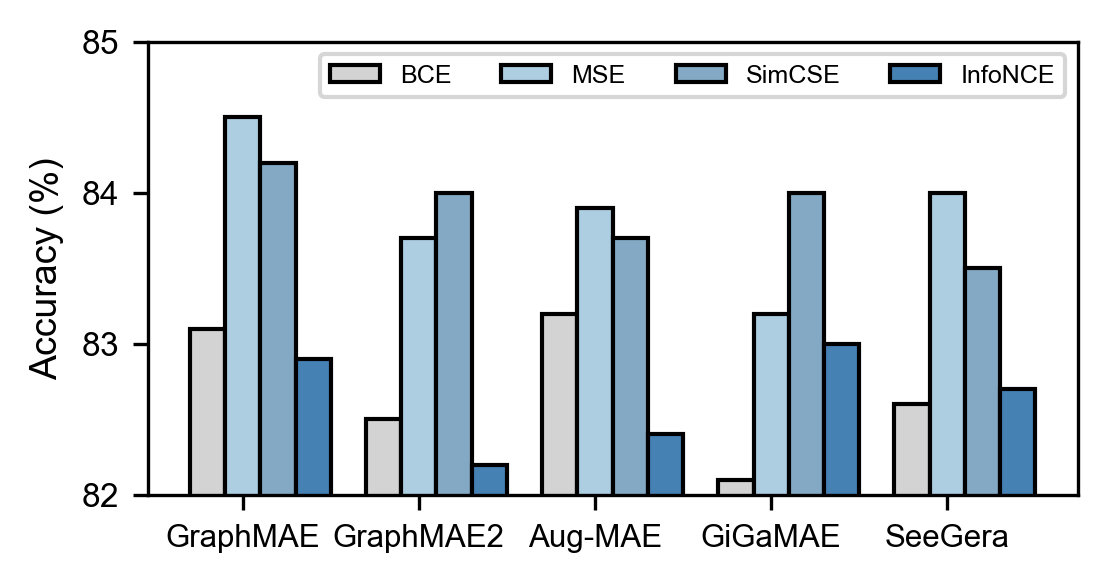}}
    \subfigure[Structure-based]{\includegraphics[width=0.48\linewidth]{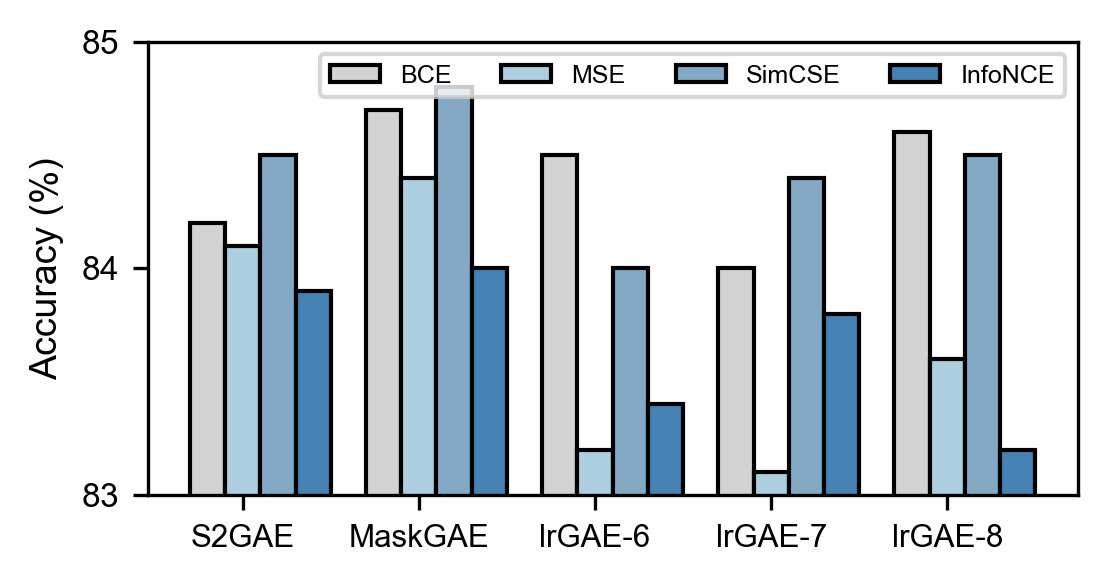}}
    \caption{Ablation on the contrastive losses.}
    \label{fig:abla_loss}
\end{figure}

\nosection{Contrastive loss.}
Next, we conduct ablation studies on the contrastive loss in GAEs, using binary cross-entropy (BCE), mean squared error (MSE), SimCSE~\cite{simcse}, and InfoNCE~\cite{infonce}. Among them, BCE and MSE are widely used for structure-based and feature-based GAEs, respectively. SimCSE and InfoNCE are two effective losses utilized in the contrastive learning literature~\cite{simcse,simclr}. 
SimCSE focuses on maximizing the similarity between augmented views of the same instance, while InfoNCE involves distinguishing positive samples from a set of negative samples. We present the results in Figure~\ref{fig:abla_loss}. The results show that although BCE and MSE are the most widely used loss functions in traditional GAEs, SimCSE shows potential as a better alternative, outperforming BCE, MSE, and InfoNCE in most GAE variants. This finding points to SimCSE as a promising direction for future research and optimization in contrastive learning within GAEs.

\section{Conclusion}
In this work, we show that a broad class of GAEs can be interpreted as implicit contrastive learners that contrast two paired subgraph views. Built upon the equivalence between GAEs and GCLs, we present \ours, a unified GAE framework that leverages the contrastive learning principles to unify existing GAE approaches. Our extensive experiments across diverse graph datasets and tasks, coupled with detailed ablation studies, have provided valuable insights into the effectiveness of contrastive views and the contributions of core components within \ours. Our work sets the foundation for a unified architecture of graph self-supervised learning, with modular and scalable GAEs and a broader understanding of the role of GAEs with different contrastive designs. Since GAEs and their masked variants are a promising research direction, \ours should be easily extended to be compatible with newly proposed approaches in the future. 
We believe \ours will have a positive impact on this emerging research domain. In the long term, we envision \ours as a core tool for advancing research in graph self-supervised learning.

\section*{Acknowledgements}
This work was supported by the New Generation Artificial Intelligence-National Science and Technology Major Project (No. 2025ZD0122701), the National Science Fund for Distinguished Young Scholars (No. 62525605).

\clearpage
\bibliographystyle{ACM-Reference-Format}
\balance
\bibliography{sample-base}

\appendix
\input{appendix}

\end{document}

%% file: math_commands.tex
\usepackage{amsmath,amsfonts,bm}

\def\eqref#1{equation~\ref{#1}}

\def\1{\bm{1}}

\DeclareMathAlphabet{\mathsfit}{\encodingdefault}{\sfdefault}{m}{sl}
\SetMathAlphabet{\mathsfit}{bold}{\encodingdefault}{\sfdefault}{bx}{n}



%% file: connection_to_gcl.tex
Motivated by recent advancements in contrastive learning~\cite{HaoChen2021Provable, zhang2022mask, maskgae}, we present a viewpoint that places both structure-based GAEs and feature-based GAEs into an (approximate) contrastive learning framework. A few additional notations will be introduced: Let $\mathcal{S}(\mathcal{G}, k)$ denote the set of all rooted subgraphs of depth $k$ with root nodes ranging over the node set $\mathcal{V}$ of the graph $\mathcal{G}$. We define an \emph{augmentation distribution} $\mathcal{A}(\cdot | s)$ to be a probability distribution that is supported on $\mathcal{S}(\mathcal{G}, k)$ conditioned on an element $s \in \mathcal{S}(\mathcal{G}, k)$. Let $p_\mathcal{G}$ be the empirical distribution of elements in $\mathcal{S}(\mathcal{G}, k)$. Following the seminal work of \cite{wang2020understanding}, we define two critical components of contrastive objectives as follows:
\begin{align}
    \mathcal{L}_\text{alignment}& = \mathbb{E}_{s^+\sim \mathcal{A}(\cdot | s), s \sim p_\mathcal{G}}\left[\ell_\text{alignment}(g_\theta(s), g_\theta(s^+))\right] \\
    \mathcal{L}_\text{uniformity}& = \mathbb{E}_{(s, s_1^-, \ldots, s_M^-)\overset{\text{i.i.d}}{\sim}p_\mathcal{G}} \left[\ell_\text{uniformity}(g_\theta(s), g_\theta(s_1^-), \ldots, g_\theta(s_M^-))\right].
\end{align}
The alignment loss $\mathcal{L}_\text{alignment}$ promotes the similarity of learned representations of positive sample pairs which are drawn according to certain augmentation mechanisms $\mathcal{A}$, while the uniformity loss $\mathcal{L}_\text{uniformity}$ encourages the diversity of learned representations to prevent collapsed solutions like constant mapping. To place GAEs in contrastive frameworks, the key step is to find suitably defined augmentation mechanisms as well as the forms of alignment and uniformity losses. We start from structure-based GAEs: Note that the first term in the right-hand side of Eq.~\ref{eq:gae1} is equivalent to a log-loss (used as an alignment loss) under the \emph{structural augmentation} $\mathcal{A}_S(\cdot | \mathcal{G}^k[v])$ defined as a uniform distribution over root-$k$ subgraphs corresponding to nodes adjacent to node $v$. However, the second term in the r.h.s of Eq.~\ref{eq:gae12} cannot be cast into exact uniformity-type losses since the negative-sampling distribution is a biased approximation of $p_\mathcal{G}$. Therefore, structure-based GAEs could be regarded as an approximated contrastive learning defined by adjacency augmentations with a biased uniformity loss, an observation which has been revealed in previous work \cite{maskgae}.\par
The case of feature-based GAEs is more complicated than that of structure-based GAEs, since the augmentation is carried out in an \emph{implicit manner}. We take inspiration from recent work \cite{zhang2022mask} by utilizing the following intuition: Two rooted subgraphs $s, s^\prime$ are considered to be a positive pair if they are likely to share the same feature of their root nodes. To present the above intuition in a slightly more formal way, denote 
\begin{align}
    \mathcal{M}_v^{(k)}(s|x) = \mathbb{P}\left[\mathcal{G}^k[v] = s\Big\vert x_v = x\right]
\end{align}
as the conditional probability measuring the likelihood of some $x \in \mathcal{S}(\mathcal{G}, k)$ being generated under root feature $x$. Then we define the following feature augmentation mechanism:
\begin{align}\label{eqn: sa}
    \mathcal{A}_\text{S}\left(\mathcal{G}^{(k)}[u]\Big\vert \mathcal{G}^{(k)}[v]\right) \propto \mathbb{E}_{x}\left[\mathcal{M}_u^{(k)}(\mathcal{G}^{(k)}[u] | x) \mathcal{M}_v^{(k)}(\mathcal{G}^{(k)}[v] | x)\right]. 
\end{align}
Consequently, we have the following justification that feature-based GAE might be approximately regarded as performing an alignment-loss minimization procedure.
\begin{lemma}
    Under mild conditions, the GAE loss (Eq.~\ref{eq:gae2}) is lower bounded by an alignment loss which is induced by the inner product:
    \begin{align}
        \mathbb{E}\left[\mathcal{L}_\text{GAE}\right] \ge \frac{1}{2} \mathbb{E}_{s^\prime \sim \mathcal{A}_F(\cdot | s), s \sim p_G}\left[ \left\langle g_\theta(s), g_\theta(s^\prime)\right \rangle \right] + \text{constant}
    \end{align}
\end{lemma}
The above lemma is a direct consequence of applying \cite[Theorem 3.4]{zhang2022mask} to our setup.\par
\nosection{Remark I: caveats of vanilla structure-based and feature-based GAEs.} Taking the perspective of (approximate) contrastive learning of GAEs would allow us to gain insights regarding their deficiency. In particular, for structure-based GAEs, directly applying Eq.~\ref{eq:gae1} would implicitly encode the information of a potentially large overlapped subgraph, thereby over-emphasizing proximity as elaborated in \cite{maskgae}. Alternatively, in feature-based GAEs, the vanilla objective Eq.~\ref{eq:gae2} contains no components that account for uniformity regularization as in standard contrastive paradigms. Consequently, the GAE loss admits trivial shortcuts as optimal solutions (the constant map) which may deteriorate learning.

\nosection{Remark II: on expositions of theoretical connections and potential implications.} 
The expositions of theoretical connections between GAEs and GCLs were largely inspired by previously established theoretical insights. To the best of our knowledge, theoretical explorations of contrastive paradigms often rely on a given augmentation distribution and derive learnability or generalization results thereafter \cite{HaoChen2021Provable, parulekar2023infonce}. However, in \ours we are more focused on the design aspects of the augmentations, with preliminary theory work presented in \cite{huang2021towards} but an extension to the field of graph learning remains highly non-trivial. Moreover, directly analyzing downstream generalization based on emerging self-supervised learning theories like \cite{lee2021predicting} is also challenging, as they typically rely on simple statistical models like linear models or topic models, which are not yet widely applicable in practical graph learning scenarios. Therefore, currently we believe a more detailed analysis tailored to the \ours framework is still beyond the scope of this paper but warrants extensive further studies.

%% file: appendix.tex
\appendix

\section{\ours framework and implementations}
\label{app:impl}

We have shown that GAEs can be viewed as generative yet implicitly contrastive models. 
In this section, we introduce \ours (\underline{l}eft-\underline{r}ight contrastive \texttt{GAE}) as a unifying formulation to analyze and instantiate contrastive view designs in GAEs, rather than as a new modeling paradigm.
Following the works in GCLs, we decompose the design space of \ours from five key dimensions: (1) augmentations, (2) contrastive views, (3) encoder/decoder networks, (4) contrastive loss, and dispensable (5) negative samples. We detail them as follows.

\subsection{Augmentations}
Graph augmentation is the first step in GCL, which generates multiple graph views from the input graph without affecting the semantic meaning~\cite{graphcl}. These views are typically created by applying certain transformations, and the goal is to help the model learn robust and generalizable representations. In GCLs, the most prevalent augmentation techniques include node dropping, edge perturbation, and attribute/feature corruption~\cite{graphcl,dropedge,dgi}.
However, as pointed out in \autoref{sec:gae_contrastive}, the contrastive viewpoint of GAEs reveals the deficiency of augmentation mechanisms like information redundancy and collapsed solutions. To mitigate these issues, a recent line of work \cite{maskgae, graphmae} has been using a simple idea of masking to improve learning performance. Specifically, in structure-based GAEs, masking a certain proportion of edges effectively reduces information redundancy \cite{maskgae}. Meanwhile, masking the root node in feature-based GAEs \cite{graphmae} can prevent trivial solutions \cite[Theorem 3.6]{zhang2022mask}. Following the design space of GCL, we mainly consider \ours with node/edge/attribute masking as augmentations.

\subsection{Encoder/decoder networks}
An encoder maps the input graph into low-dimensional representations and is typically defined as a GNN network. Basically, the receptive fields of the encoder are determined by the depth of the GNN network. Meanwhile, the decoder network is regarded as a task-specific `adapter', which maps augmented representations to another latent space where the contrastive loss is calculated for different pretext tasks, such as graph reconstruction.
In most cases, GAEs employ an asymmetric design, where the decoder network is implemented as an MLP, although a GNN can also be used to enhance decoding and expand the receptive fields. However, using a GNN as the decoder may additionally introduce the oversmoothing issue.
Multiple encoders and decoders can be adopted to learn diverse representations in different ways.
Similar to GCLs, \ours does not apply any constraint on the encoder/decoder architecture.

\begin{table}[h]
  \label{tab:complexity}
  \centering
  \small
  \caption{Time and space complexity of \ours framework. $|E|$ is the number of edges and $|V|$ is the number of nodes, $L$ is the number of layers, $d$ is the number of features. For simplicity, we assume the number of features is fixed across all layers. For the mini-batch training setting, $B$ is the batch size and $r$ is the number of sampled neighbors per node.}
  \resizebox{\linewidth}{!}
  {
    \begin{threeparttable}
      \begin{tabular}{lcc}
        \toprule
                          & \textbf{Full-batch training}                     & \textbf{Mini-batch training}       \\ \midrule
        Time complexity   & $\mathcal{O}(L|\mathcal{E}|d+L|\mathcal{V}|d^2)$ & $\mathcal{O}(r^L|\mathcal{V}|d^2)$ \\
        Memory complexity & $\mathcal{O}(L|\mathcal{V}|d+Ld^2)$              & $\mathcal{O}(Br^Ld+Ld^2)$          \\ \bottomrule
      \end{tabular}
    \end{threeparttable}
  }
\end{table}

\begin{figure*}
  \centering
  \includegraphics[width=0.9\linewidth]{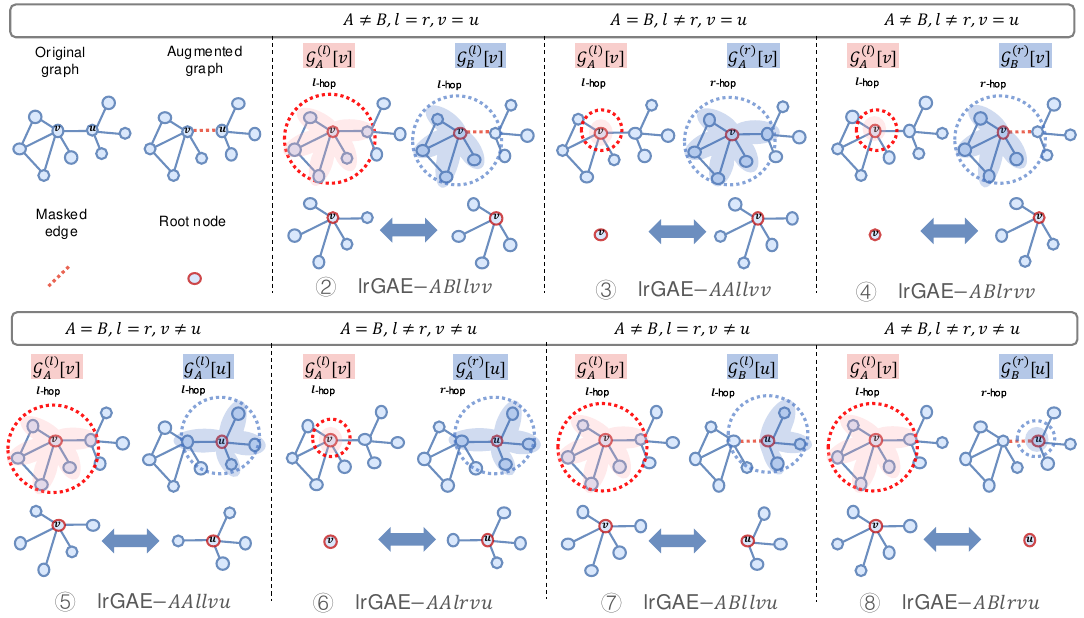}
  \caption{Illustration of seven possible cases of \ours. We vary the graph views ($A$ and $B$), receptive fields ($l$ and $r$), as well as node pairs ($v$ and $u$) to implement different variants of \ours with different contrastive views.}
  \label{fig:cases}
\end{figure*}

\subsection{Contrastive loss}
Contrastive loss or objective is designed to train models to distinguish between similar (positive) and dissimilar (negative) pairs of graph views according to different pretext tasks. The main idea in vanilla GCLs is to bring the representations of augmented views of the graph closer together while pushing the representations of different graphs further apart.
By effectively contrasting positive and negative pairs, the model learns to capture the essential structural and feature information of the graphs, leading to improved performance on various downstream tasks.
For GAEs, whose core idea is to perform reconstruction over graph structure or node features, the learning objectives involve binary cross-entropy (BCE) loss or regression loss mean squared error (MSE), respectively.
As discussed in \autoref{sec:gae_contrastive}, both losses follow the same contrastive learning philosophy. As we have established a strong connection between GAEs and GCLs, we can further advance GAEs with various well-established contrastive learning objectives beyond simple BCE or MSE in \ours.

\subsection{Negative samples}
Negative samples are instances that are dissimilar to the positive samples in contrastive learning, which act as a reference for the learning model to distinguish between meaningful patterns and random noise.
Conventional GCLs largely rely on negative samples to learn meaningful representations and capture important patterns in the data. However, this approach comes with a trade-off in learning efficiency, and the quality of negative samples is not always guaranteed.
Since graph structure reconstruction in GAEs is inherently tackled as a binary link classification task, negative links or edges are necessary to prevent model collapse and encourage learning meaningful representations. In contrast, feature reconstruction focuses on recovering node-level or graph-level features without the need for negative samples since the task revolves around reproducing the original feature representations.
Recent advances have shown that GCLs can also be effective even without explicit negative samples~\cite{bgrl,cca_ssg,sgcl}. This opens up the possibility of eliminating negative samples in \ours by properly incorporating several techniques from GCL such as stop-gradient or asymmetric networks.

\subsection{Algorithmic description.}
To make the design space in Eq.~\ref{eq:gae_gcl} operational, we now describe a generic training procedure for \ours that separates \emph{view construction} from \emph{objective instantiation}. Specifically, given an input graph, \ours first generates paired subgraph views by selecting graph augmentations, receptive fields, and node pair sets, and then applies an encoder/decoder with a task-dependent contrastive loss. This framework allows different GAE variants and tasks to share the same algorithmic skeleton while differing only in a few plug-in choices. We provide PyTorch~\cite{pytorch}-style pseudocode for \ours in Algorithm~\ref{algo:lrgae}.

\begin{figure}[h]
  \begin{minipage}{\linewidth}
    \vspace{-6mm}
    \begin{algorithm}[H]
      \caption{PyTorch~\cite{pytorch} style pseudocode for \ours.}
      \label{algo:lrgae}
      \begin{algorithmic}[0]
        \State \PyComment{g:\ input graph}
        \State \PyComment{gA, gB:\ graph views}
        \State \PyComment{l, r:\ receptive fields or hops}
        \State \PyComment{v, u:\ contrastive node pairs}
        \State
        \State \PyComment{augmentation}
        \State \PyCode{gA, gB = \pink{augmentation}(g)}
        \State
        \State \PyComment{encoding}
        \State \PyCode{left = \pink{encoder}(gA)[l]}
        \State \PyCode{right = \pink{encoder}(gB)[r]}
        \State
        \State \PyComment{decoding}
        \State \PyCode{viewV = \pink{decoder}(left[v]})
        \State \PyCode{viewU = \pink{decoder}(right[u]})
        \State
        \State \PyComment{contrasting}
        \State \PyCode{loss = \pink{loss\_fn}(viewV, viewU)}
      \end{algorithmic}
    \end{algorithm}
  \end{minipage}
  \vspace{-3mm}
\end{figure}

\subsection{\ours implementations}
To showcase the versatility and broad applicability of \ours, we list seven variants of \ours with different contrastive views below:
\begin{itemize}
  \item \underline{\ours~\circled{2}-$ABllvv$}: This variant can be simply implemented with the same architecture as na\"ive GCLs~\cite{grace,graphcl}, which contrast two augmentation views ($\mathcal{G}_A$ and $\mathcal{G}_B$) for each node ($v$) in its $l$-hop neighborhood (i.e., receptive field).
  \item \underline{\ours~\circled{3}-$AAlrvv$}: This is a variant of \ours~\circled{2}-$ABllvv$ that contrasts the node representations from different layers of the encoder, i.e., the receptive fields ($l$ and $r$), which shares a similar philosophy of local-to-global graph contrastive learning~\cite{dgi}. The most typical example is GAE~\cite{gae} with feature reconstruction as its learning objective (i.e., GAE$_f$).
  \item \underline{\ours~\circled{4}-$ABlrvv$}: This variant is the core implementation of GraphMAE~\cite{graphmae}. By combining the success of GCLs and GAEs, we are able to develop this variant that incorporates different graph views ($\mathcal{G}_A$ and $\mathcal{G}_B$) and receptive fields ($l$ and $r$) of a node $v$.
  \item \underline{\ours~\circled{5}-$AAllvu$}: This is the learning paradigm of vanilla GAE and MaskGAE~\cite{maskgae}, which contrasts the two subgraph pairs of nodes associated with an edge $(v, u)$. Note that the graph view $\mathcal{G}_A$ can be the original graph $\mathcal{G}$ or the augmented/masked graph, leading to the implementations of GAE or MaskGAE.
  \item \underline{\ours~\circled{6}-$AAlrvu$, \ours~\circled{7}-$ABllvu$, and \ours~\circled{8}-$ABlrvu$}: These variants do not have established implementations in prior work; we implement them within \ours. Following the architecture of \ours~\circled{5}-$AAllvu$, we can vary the receptive fields ($l$ and $r$), augmentation views ($\mathcal{G}_A$ and $\mathcal{G}_B$), or even both to perform graph contrastive learning.
\end{itemize}
As shown in Table~\ref{tab:cases}, \ours~\circled{2}\circled{3}\circled{4}\circled{5} have implementations proposed in previous works~\cite{grace,graphcl,gae,graphmae,maskgae}, and we mainly provide the empirical results of \ours~\circled{6}/\circled{7}/\circled{8} in our experiments. The above variants cover the possible implementations of GAEs, further demonstrating the flexibility and adaptability of \ours. This allows researchers to choose and customize the appropriate contrastive views based on their specific tasks and objectives. We present the illustration of the seven variants of \ours in Figure~\ref{fig:cases}.

\section{Discussions}

\begin{table*}[t]
    \centering
    \small
    \caption{Comparison of different (masked) GAEs that fall into our contrastive framework \ours. GCL is also included in the table for reference. Here $k$ is the depth/receptive field of the encoder/decoder networks. BCE: binary cross-entropy; MSE: mean squared error; SCE: scaled cosine error.}\label{tab:comparison}
    \begin{threeparttable}
        {
            {
                    \begin{tabular}{lcccc}
                        \toprule
                                                            & \textbf{Augmentations}       & \textbf{Contrastive views}                                                                                              & \textbf{Contrastive Loss} & \textbf{Negative samples} \\
                        \midrule
                        \textbf{GCL}                        & Edge/feature corruption, ... & $\leftg{{\mathcal{G}^{(k)}_A[v]}}\leftrightarrow\rightg{{\mathcal{G}^{(k)}_B[v]}}$                                      & InfoNCE, SimCSE, ...      & \cmark$^\dagger$          \\
                        \midrule
                        \textbf{GAE$_f$}~\cite{gae}         & -                            & $\leftg{{\mathcal{G}^{(k)}[v]}}\leftrightarrow\rightg{{\mathcal{G}^{(0)}[v]}}$                                          & MSE                       & \xmark                    \\
                        \textbf{GAE}~\cite{gae}             & -                            & $\leftg{{\mathcal{G}^{(k)}[v]}}\leftrightarrow\rightg{{\mathcal{G}^{(k)}[u]}}$                                          & BCE                       & \cmark                    \\
                        \textbf{MaskGAE}~\cite{maskgae}     & Edge/path masking            & $\leftg{{\mathcal{G}^{(k)}_A[v]}}\leftrightarrow\rightg{{\mathcal{G}^{(k)}_A[u]}}$                                      & BCE                       & \cmark                    \\
                        \textbf{S2GAE}~\cite{s2gae}         & Directed edge masking        & $\leftg{{\mathcal{G}^{(k)}_A[v]}}\leftrightarrow\rightg{{\mathcal{G}^{(k)}_A[u]}}$                                      & BCE                       & \cmark                    \\
                        \textbf{GraphMAE}~\cite{graphmae}   & Feature masking              & $\leftg{{\mathcal{G}^{(k)}_A[v]}}\leftrightarrow\rightg{{\mathcal{G}^{(0)}_B[v]}}$                                      & MSE/SCE                   & \xmark                    \\
                        \textbf{GraphMAE2}~\cite{graphmae2} & Feature masking              & $\leftg{{\mathcal{G}^{(k)}_A[v]}}\leftrightarrow\rightg{{\mathcal{G}^{(0)}_B[v]}} \cup \rightg{{\mathcal{G}^{(k)}[v]}}$ & MSE/SCE                   & \xmark                    \\
                        \textbf{AUG-MAE}~\cite{augmae}      & Adaptive feature masking     & $\leftg{{\mathcal{G}^{(k)}_A[v]}}\leftrightarrow\rightg{{\mathcal{G}^{(0)}_B[v]}}$                                      & MSE/SCE                   & \xmark                    \\
                        \textbf{GiGaMAE}~\cite{gigamae}     & Edge \& feature masking      & $\leftg{{\mathcal{G}^{(k)}_A[v]}}\leftrightarrow\rightg{{\mathcal{G}^{(0)}_B[v]}}$                                      & InfoNCE                   & \xmark                    \\
                        \textbf{SeeGera}~\cite{seegera}     & Edge \& feature masking      & $\leftg{{\mathcal{G}^{(k)}_A[v]}}\leftrightarrow\rightg{{\mathcal{G}^{(0)}_B[v]}  \cup {\mathcal{G}^{(k)}_B[u]} }$      & BCE/MSE/SCE               & \xmark                    \\
                        \bottomrule
                    \end{tabular}
                }
            \begin{tablenotes}
                \item[$\dagger$]\scriptsize Note that there are a set of works in GCLs that do not require negative samples~\cite{bgrl,cca_ssg,sgcl}.
            \end{tablenotes}
        }
    \end{threeparttable}
    \vspace{-3mm}
\end{table*}

\nosection{Time and space complexity.}
We briefly discuss the time and space complexity of our proposed \ours framework. \ours is a standard GAE framework consisting of one encoder and one decoder network.
As decoder networks are typically simple feed-forward networks (e.g., MLPs) involved with dense matrix computation, the major bottleneck arises from the message passing and aggregation of encoder networks (e.g., GCN~\cite{kipf2016semi}). In particular, for an $L$ layer encoder network, the time complexity is related to the graph size and the dimension of features and hidden representations, about $\mathcal{O}(L|\mathcal{E}|d+L|\mathcal{V}|d^2)$.
By incorporating mini-batch training~\cite{hamilton2017inductive}, the time complexity for each sampled subgraph can be reduced to $\mathcal{O}(r^L|\mathcal{V}|d^2)$, where $r$ represents the neighborhood size shared by each hop.
As for the space complexity, the major bottleneck also lies in the graph encoder network, which is $\mathcal{O}(L|\mathcal{V}|d+Ld^2)$ and $\mathcal{O}(Br^Ld+Ld^2)$ for full-batch and mini-batch training, respectively. Here, $B$ denotes the batch size. Overall, the time and space complexity of \ours framework are guaranteed and can easily scale to larger graphs, showcasing its scalability and generality.

\nosection{Adapting current GAEs to \ours framework.}
Since we have established the close connections between GAEs and GCLs, we are now able to relate popular GAEs to our proposed \ours framework. Table~\ref{tab:comparison} summarizes several advanced GAEs as well as conventional GCLs within our proposed contrastive framework across four dimensions. We omit the encoder/decoder networks here as most methods share a similar network architecture. GAEs are tailored by their asymmetric contrastive views design, either on graph views or receptive fields. In addition, most existing GAEs follow the conventional learning paradigms, i.e., regression loss (MSE or SCE~\cite{graphmae}) for feature-based reconstruction, while classification loss is used for structure-based reconstruction. Until recently, only GiGaMAE has attempted to adapt other GCL losses (e.g., InfoNCE~\cite{Tschannen2020On,simclr}) to GAEs, leaving the potential of GAEs with more powerful GCL techniques largely unexplored.

\nosection{Comparison with existing work.}
Recent work has begun to explicitly analyze the relationship between GAEs and GCL. In particular, AUG-MAE~\cite{augmae} is closely related to our motivation, as it connects masked GAEs to GCL. It shows that the node-level feature reconstruction loss in GraphMAE admits a lower bound in terms of a context-level alignment objective, suggesting that masked GAEs can implicitly align paired ego-graph views~\cite{augmae}.
Our contribution goes beyond this conclusion in both scope and form. First, AUG-MAE’s theory is tailored to the feature-masked, node-wise reconstruction setting with ego-graph contexts, so its bridge to GCL is established for that specific masked objective. In contrast, our theory abstracts GAEs through paired subgraph views and shows that implicit contrastive learning serves as a general organizing principle across a broader family of GAE objectives and view constructions, with masked feature reconstruction as one special case. Second, we operationalize this broader view by introducing lrGAE as an explicit design space for contrastive view construction in GAEs, enabling systematic comparison and motivating new asymmetric variants that are not covered by prior analyses. Empirically, this framework allows us to study how different view pairings and receptive-field choices affect performance across tasks, rather than refining a single masking strategy.

\nosection{Limitations and outlook.} This work primarily presents initial benchmarks and baselines for GAEs, with a main focus on masked autoencoding-based methods.
Our work might potentially suffer from some limitations: (i) Graph augmentation, such as masking, is a core operation of \ours. However, similar to existing augmentation-based GCL methods, performing masking on the graph structure could harm the semantic meaning of some graphs, such as biochemical molecules~\cite{DBLP:conf/ijcai/ZhaoLHLZ21}. (ii) The theoretical results of \ours are mainly based on the homophily assumption, a basic assumption in the literature of graph-based representation learning. However, such an assumption may not always hold in heterophilic graphs, where the labels of linked nodes are likely to differ. This could potentially limit the applicability of \ours on heterophilic graphs.
Our current analysis is based solely on existing datasets; hence, we plan to enhance the scope of \ours by acquiring and incorporating more diverse datasets and domains.
Furthermore, while this paper primarily conducts an empirical investigation on GAEs, advancing the theoretical framework of \ours will be crucial for its development and understanding.

\begin{table}[h]
  \centering
  \small
  \caption{Statistics of datasets for three fundamental graph learning tasks. \red{N}: node classification; \blue{L}: link prediction; \orange{C}: graph clustering.}\label{tab:dataset}
  {\begin{tabular}{lccccc}
      \toprule
      {\textbf{Dataset}} & {\#Nodes} & {\#Edges} & {\#Features} & {\#Classes} & {Tasks}                     \\
      \midrule
      \textbf{Cora}      & 2,708     & 10,556     & 1,433        & 7           & \red{N}/\blue{L}/\orange{C} \\
      \textbf{CiteSeer}  & 3,327     & 9,104      & 3,703        & 6           & \red{N}/\blue{L}/\orange{C} \\
      \textbf{PubMed}    & 19,717    & 88,648     & 500          & 3           & \red{N}/\blue{L}/\orange{C} \\
      \textbf{Photo}     & 7,650     & 238,162    & 745          & 8           & \red{N}/\orange{C}          \\
      \textbf{Computer}  & 13,752    & 491,722    & 767          & 10          & \red{N}/\orange{C}          \\
      \textbf{CS}        & 18,333    & 163,788    & 6,805        & 15          & \red{N}/\orange{C}          \\
      \textbf{Physics}   & 34,493    & 495,924    & 8,415        & 5           & \red{N}/\orange{C}          \\
      \midrule
      \multicolumn{6}{c}{Open Graph Benchmark Datasets} \\
      \midrule
      \textbf{arXiv}     & 169,343   & 2,315,598  & 128          & 40          & \red{N}                     \\
      \textbf{MAG}       & 1,939,743 & 21,111,007 & 128          & 349         & \red{N}                     \\
      \textbf{Products}  & 2,449,029 & 61,859,140 & 100          & 47          & \red{N}                     \\
      \textbf{ddi}  & 4,267 & 1,334,889 & -          & -          & \blue{L}                     \\
      \textbf{collab}  & 235,868 & 1,285,465 & 128          &    -       & \blue{L}                     \\
      \textbf{ppa}  & 576,289 & 30,326,273 & -          & -          & \blue{L}                     \\
      \bottomrule
    \end{tabular}
  }
\end{table}

\begin{table}[h]
  \centering
  \small
  \caption{Statistics of datasets for graph classification task (\mygreen{G}).}\label{tab:graph_dataset}
  \resizebox{\linewidth}{!}
  {\begin{tabular}{lcccccc}
      \toprule
      \textbf{Dataset}  & Avg. \#nodes & Avg. \#edges & \#Graphs & \#Features & \#Classes & Task        \\
      \midrule
      \textbf{IMDB-B}   & 19.8         & 193.1        & 1,000    & 0          & 2         & \mygreen{G} \\
      \textbf{IMDB-M}   & 13.0         & 131.9        & 1,500    & 0          & 3         & \mygreen{G} \\
      \textbf{PROTEINS} & 39.1         & 145.6        & 1,113    & 4          & 2         & \mygreen{G} \\
      \textbf{COLLAB}   & 74.5         & 4914.4       & 5,000    & 0          & 3         & \mygreen{G} \\
      \textbf{MUTAG}    & 17.9         & 39.6         & 188      & 7          & 2         & \mygreen{G} \\
      \textbf{REDDIT-B} & 429.6        & 995.5        & 2,000    & 0          & 2         & \mygreen{G} \\
      \textbf{NCI1}     & 29.9         & 64.6         & 4,110    & 37         & 2         & \mygreen{G} \\
      \bottomrule
    \end{tabular}
  }
\end{table}

\begin{table}[h]
  \centering
  \small
  \caption{Statistics of datasets for heterogeneous node classification task (\pink{HN}).}\label{tab:hg_dataset}
  \resizebox{\linewidth}{!}{
    \begin{tabular}{lcccccc}
      \toprule
      \textbf{Dataset} & \#Nodes & \#Edges & \#Node types & \#Edge types & \#Classes & Task      \\
      \midrule
      \textbf{DBLP}    & 26,128  & 239,566 & 4            & 6            & 4         & \pink{HN} \\
      \textbf{ACM}     & 10,942  & 547,872 & 4            & 8            & 3         & \pink{HN} \\
      \textbf{IMDB}    & 21,420  & 86,642  & 4            & 6            & 5         & \pink{HN} \\
      \bottomrule
    \end{tabular}
  }
\end{table}

\section{Reproducibility}
\label{appendix:setting}

All of \ours's experimental results are highly reproducible. We provide more detailed information on the following aspects to ensure the reproducibility of the experiments.

\nosection{Datasets.}
We conduct experiments on seven graph benchmark datasets, including three citation networks, i.e., Cora, CiteSeer, and PubMed~\cite{sen2008collective}, two Amazon co-purchase graphs, i.e., Photo and Computer~\cite{shchur2018pitfalls}, and two co-author graphs, i.e., CS and Physics~\cite{shchur2018pitfalls}.
The above datasets are used for the tasks of node classification, link prediction, and graph clustering.
For the graph classification task, we perform experiments on the following seven datasets: IMDB-B, IMDB-M, PROTEINS, COLLAB, MUTAG, REDDIT-B, and NCI1~\cite{tudataset}. Each dataset consists of a set of graphs, with each graph associated with a corresponding label.
We also incorporate three heterogeneous graph datasets, i.e., DBLP, ACM, and IMDB~\cite{lv2021we,hetero2net}, for experiments on heterogeneous node classification to showcase the generality and versatility of our \ours framework.
All datasets used throughout the experiments are publicly available at PyTorch Geometric~\cite{pyg}.
Detailed information about the datasets is summarized in Table~\ref{tab:dataset}, Table~\ref{tab:graph_dataset} and Table~\ref{tab:hg_dataset}.

\begin{table*}[t]
    \centering
    \small
    \caption{Graph clustering NMI (\%) on seven benchmark datasets. \texttt{OOM}: out-of-memory on an NVIDIA 3090ti GPU with 24GB memory.}
    \label{tab:graph_cluster}
    \begin{threeparttable}{
            {
                    \begin{tabular}{llccccccc}
                        \toprule
                                                                                         &                                      & \textbf{Cora}                             & \textbf{CiteSeer}                           & \textbf{PubMed}                    & \textbf{Photo}                     & \textbf{Computer}                        & \textbf{CS}                                 & \textbf{Physics}                          \\

                        \midrule
                        \multirow{7}{*}{\rotatebox{0}{\thead{\textbf{Standard GCL}}}}
                                                                                         & DGI~\cite{dgi}                       & 55.1$_{\pm2.0}$                           & 43.5$_{\pm1.7}$                             & 31.0$_{\pm0.6}$                    & 33.7$_{\pm1.2}$                    & 42.0$_{\pm1.3}$                           & 68.5$_{\pm0.7}$                             & 60.2$_{\pm0.8}$                           \\
                                                                                         & GRACE~\cite{grace}                   & 56.0$_{\pm1.4}$                           & 39.9$_{\pm1.5}$                             & 32.5$_{\pm0.6}$                    & 53.5$_{\pm1.5}$                    & 53.8$_{\pm1.2}$                           & 71.0$_{\pm0.6}$                             & 62.8$_{\pm0.7}$                           \\
                                                                                         & BGRL~\cite{bgrl}                     & 57.4$_{\pm1.2}$                           & 42.2$_{\pm0.8}$                             & \best{34.0$_{\pm0.5}$}             & 50.1$_{\pm1.6}$                    & 44.1$_{\pm1.4}$                           & 72.6$_{\pm0.5}$                             & 64.0$_{\pm0.6}$                           \\
                                                                                         & CCA-SSG~\cite{cca_ssg}               & 56.6$_{\pm0.9}$                           & 41.8$_{\pm1.3}$                             & 33.2$_{\pm0.6}$                    & 51.2$_{\pm1.4}$                    & 46.8$_{\pm1.3}$                           & 73.4$_{\pm0.5}$                             & 65.1$_{\pm0.6}$                           \\
                                                                                         & GGD~\cite{ggd}                       & 55.9$_{\pm1.2}$                           & 40.5$_{\pm1.2}$                             & 32.0$_{\pm0.7}$                    & 48.4$_{\pm1.5}$                    & 45.0$_{\pm1.4}$                           & 71.9$_{\pm0.6}$                             & 63.2$_{\pm0.7}$                           \\
                                                                                         & S3GCL~\cite{s3gcl}                   & 55.8$_{\pm1.5}$                           & 44.0$_{\pm0.8}$                             & 33.5$_{\pm0.6}$                    & 50.9$_{\pm1.4}$                    & 47.2$_{\pm1.2}$                           & 74.1$_{\pm0.4}$                             & 66.0$_{\pm0.6}$                           \\
                                                                                         & EPAGCL~\cite{epagcl}                 & 56.9$_{\pm2.4}$                           & 41.2$_{\pm1.2}$                             & 33.8$_{\pm0.5}$                    & 49.0$_{\pm1.6}$                    & 46.0$_{\pm1.4}$                           & 72.3$_{\pm0.6}$                             & 64.5$_{\pm0.7}$                           \\

                        \midrule
                        \multirow{7}{*}{\rotatebox{0}{\thead{\textbf{Feature-based}}}}   & GAE$_f$~\cite{gae}                   & 13.1$_{\pm2.5}$                           & 2.6$_{\pm1.5}$                              & 4.0$_{\pm0.5}$                     & 27.1$_{\pm3.5}$                    & 10.3$_{\pm2.5}$                           & 20.2$_{\pm0.6}$                             & 50.3$_{\pm1.9}$                           \\
                                                                                         & GraphMAE~\cite{graphmae}             & 54.3$_{\pm2.1}$                           & \second{44.9$_{\pm1.7}$}                    & {33.2$_{\pm0.4}$}                  & 66.9$_{\pm3.2}$                    & \second{56.8$_{\pm2.4}$}                  & 76.2$_{\pm0.8}$                             & 64.6$_{\pm1.5}$                           \\
                                                                                         & GraphMAE2~\cite{graphmae2}           & 54.1$_{\pm2.4}$                           & \best{46.7$_{\pm1.0}$}                      & \second{33.9$_{\pm0.8}$}           & 66.8$_{\pm3.7}$                    & 54.8$_{\pm1.5}$                           & 76.2$_{\pm0.4}$                             & 65.1$_{\pm1.2}$                           \\
                                                                                         & AUG-MAE~\cite{augmae}                & 57.6$_{\pm1.8}$                           & 44.6$_{\pm1.2}$                             & 33.5$_{\pm1.0}$                    & \best{70.8$_{\pm2.3}$}             & 54.6$_{\pm1.3}$                           & \texttt{OOM}                                & \texttt{OOM}                              \\
                                                                                         & GiGaMAE~\cite{gigamae}               & 55.7$_{\pm1.9}$                           & 37.0$_{\pm1.6}$                             & 34.0$_{\pm0.8}$                    & \second{69.6$_{\pm2.9}$}           & 56.4$_{\pm1.7}$                           & 74.7$_{\pm0.3}$                             & \texttt{OOM}                              \\
                                                                                         & SeeGera~\cite{seegera}               & 56.1$_{\pm1.2}$                           & 39.3$_{\pm1.9}$                             & 30.1$_{\pm0.5}$                    & {68.1$_{\pm1.4}$}                  & 55.2$_{\pm1.8}$                           & 76.7$_{\pm0.4}$                             & \texttt{OOM}                              \\
                        \midrule
                        \multirow{6}{*}{\rotatebox{0}{\thead{\textbf{Structure-based}}}} & GAE~\cite{gae}                       & 51.8$_{\pm2.4}$                           & 33.0$_{\pm1.8}$                             & 24.8$_{\pm0.9}$                    & 54.4$_{\pm2.4}$                    & 48.2$_{\pm3.1}$                           & 72.8$_{\pm0.9}$                             & 61.0$_{\pm0.6}$                           \\
                                                                                         & S2GAE~\cite{s2gae}                   & 55.6$_{\pm2.6}$                           & 32.8$_{\pm2.0}$                             & 8.9$_{\pm1.1}$                     & 59.3$_{\pm4.1}$                    & 50.1$_{\pm3.3}$                           & 71.5$_{\pm1.1}$                             & 67.9$_{\pm0.5}$                           \\
                                                                                         & MaskGAE~\cite{maskgae}               & \second{58.0$_{\pm2.4}$}                  & 43.8$_{\pm1.4}$                             & 28.6$_{\pm1.2}$                    & 58.2$_{\pm3.5}$                    & 56.3$_{\pm2.9}$                           & 76.8$_{\pm0.8}$                             & \second{74.9$_{\pm0.9}$}                  \\
                                                                                         & \cellcolor{gray!10}\ours~\circled{6} & \cellcolor{gray!10}56.4$_{\pm2.1}$        & \cellcolor{gray!10}43.7$_{\pm1.6}$          & \cellcolor{gray!10}28.0$_{\pm0.8}$ & \cellcolor{gray!10}59.1$_{\pm3.7}$ & \cellcolor{gray!10}51.8$_{\pm1.9}$        & \cellcolor{gray!10}77.0$_{\pm0.5}$          & \cellcolor{gray!10}72.1$_{\pm0.8}$        \\
                                                                                         & \cellcolor{gray!10}\ours~\circled{7} & \cellcolor{gray!10}\best{59.0$_{\pm2.6}$} & \cellcolor{gray!10}\second{44.9$_{\pm1.8}$} & \cellcolor{gray!10}27.3$_{\pm0.9}$ & \cellcolor{gray!10}64.4$_{\pm3.8}$ & \cellcolor{gray!10}\best{57.0$_{\pm2.2}$} & \cellcolor{gray!10}\second{77.2$_{\pm0.7}$} & \cellcolor{gray!10}69.0$_{\pm0.4}$        \\
                                                                                         & \cellcolor{gray!10}\ours~\circled{8} & \cellcolor{gray!10}57.3$_{\pm2.7}$        & \cellcolor{gray!10}44.0$_{\pm1.8}$          & \cellcolor{gray!10}30.7$_{\pm0.5}$ & \cellcolor{gray!10}53.8$_{\pm3.2}$ & \cellcolor{gray!10}50.2$_{\pm2.4}$        & \cellcolor{gray!10}\best{77.3$_{\pm0.5}$}   & \cellcolor{gray!10}\best{76.0$_{\pm0.5}$} \\
                        \bottomrule
                    \end{tabular}
                }
        }
    \end{threeparttable}
\end{table*}

\begin{table*}[t]
  \centering
  \small
  \caption{Graph classification accuracy (\%) on seven benchmark datasets.}  \label{tab:graph_clas}
  \begin{threeparttable}{
      {
          \begin{tabular}{llccccccc}
            \toprule
                  &            & \textbf{IMDB-B}                  & \textbf{IMDB-M}          & \textbf{PROTEINS}        & \textbf{COLLAB}          & \textbf{MUTAG}           & \textbf{REDDIT-B}        & \textbf{NCI1}            \\

            \midrule
\multirow{7}{*}{\rotatebox{0}{\thead{\textbf{Standard GCL}}}}
& DGI~\cite{dgi}         & 71.1$_{\pm0.5}$ & 49.4$_{\pm0.9}$ & 74.3$_{\pm2.0}$ & 70.7$_{\pm1.4}$ & 83.2$_{\pm1.3}$ & 79.8$_{\pm1.0}$ & 76.2$_{\pm1.8}$ \\
& GRACE~\cite{grace}     & 73.0$_{\pm0.9}$ & 50.9$_{\pm0.5}$ & 75.4$_{\pm0.3}$ & 79.6$_{\pm1.3}$ & 89.0$_{\pm1.1}$ & 88.5$_{\pm1.4}$ & 77.2$_{\pm1.5}$ \\
& BGRL~\cite{bgrl}       & 73.2$_{\pm3.1}$ & 52.2$_{\pm0.8}$ & 76.9$_{\pm0.4}$ & 81.5$_{\pm0.7}$ & 89.4$_{\pm1.0}$ & 87.3$_{\pm1.7}$ & 79.4$_{\pm0.6}$ \\
& CCA-SSG~\cite{cca_ssg} & 72.2$_{\pm0.5}$ & 50.2$_{\pm0.3}$ & 76.8$_{\pm1.2}$ & 79.4$_{\pm0.6}$ & 89.5$_{\pm0.7}$ & 86.7$_{\pm0.8}$ & 78.8$_{\pm0.5}$ \\
& GGD~\cite{ggd}         & 71.9$_{\pm0.4}$ & 50.4$_{\pm0.8}$ & 75.5$_{\pm0.8}$ & 76.3$_{\pm1.2}$ & 85.4$_{\pm0.3}$ & 84.5$_{\pm0.9}$ & 76.5$_{\pm0.3}$ \\
& S3GCL~\cite{s3gcl}     & 72.4$_{\pm0.5}$ & 51.5$_{\pm0.4}$ & 76.6$_{\pm0.9}$ & 80.7$_{\pm1.4}$ & 86.5$_{\pm0.9}$ & 86.2$_{\pm0.8}$ & 79.6$_{\pm0.5}$ \\
& EPAGCL~\cite{epagcl}   & 72.2$_{\pm0.7}$ & 52.0$_{\pm0.7}$ & 76.9$_{\pm0.4}$ & 79.4$_{\pm1.3}$ & 89.7$_{\pm1.3}$ & 86.0$_{\pm1.6}$ & 80.5$_{\pm0.4}$ \\
                  
            \midrule
            \multirow{7}{*}{\rotatebox{0}{\thead{\textbf{Feature-based}}}} & GAE$_f$~\cite{gae}           & 74.4$_{\pm0.8}$                  & 52.5$_{\pm0.7}$          & 75.3$_{\pm1.2}$          & 76.9$_{\pm1.5}$          & 87.0$_{\pm1.2}$          & 72.8$_{\pm2.5}$          & 74.5$_{\pm1.3}$          \\
            & GraphMAE~\cite{graphmae}          & 75.0$_{\pm0.6}$                  & 52.1$_{\pm0.4}$          & 75.8$_{\pm0.7}$          & \best{82.7$_{\pm1.0}$}   & 89.3$_{\pm1.1}$          & \best{88.8$_{\pm2.8}$}   & 80.1$_{\pm1.0}$          \\
            & GraphMAE2~\cite{graphmae2}         & \second{75.5$_{\pm0.7}$}         & \best{52.7$_{\pm0.6}$}   & 75.4$_{\pm0.5}$          & 81.7$_{\pm0.8}$          & 89.5$_{\pm1.5}$          & 88.6$_{\pm2.3}$          & \best{82.3$_{\pm0.9}$}   \\
            & AUG-MAE~\cite{augmae}           & \best{75.6$_{\pm1.1}$}           & 52.2$_{\pm1.0}$          & 73.5$_{\pm0.8}$          & 79.9$_{\pm0.4}$          & 89.8$_{\pm1.3}$          & 88.3$_{\pm3.1}$          & 78.8$_{\pm1.2}$          \\
            & GiGaMAE~\cite{gigamae}           & \multicolumn{6}{c}{\texttt{N/A}}                                                                                                                                                                   \\
            & SeeGera~\cite{seegera}           & {72.6$_{\pm1.3}$}           & 51.4$_{\pm1.3}$          & 74.6$_{\pm0.4}$          & 80.8$_{\pm0.6}$          & 88.9$_{\pm1.2}$          & 88.5$_{\pm1.1}$          & 77.6$_{\pm1.2}$          \\
            \midrule
            \multirow{6}{*}{\rotatebox{0}{\thead{\textbf{Structure-based}}}} & GAE~\cite{gae}               & 75.1$_{\pm0.7}$                  & 51.5$_{\pm1.5}$          & 76.6$_{\pm0.9}$          & 80.1$_{\pm0.6}$          & 89.5$_{\pm1.5}$          & 82.5$_{\pm2.5}$          & 73.8$_{\pm1.7}$          \\
            & S2GAE~\cite{s2gae}             & 73.6$_{\pm0.6}$                  & 52.5$_{\pm0.9}$          & 76.0$_{\pm0.3}$          & 82.2$_{\pm0.4}$          & 85.7$_{\pm0.8}$          & \second{89.4$_{\pm2.7}$} & 77.2$_{\pm0.8}$          \\
            & MaskGAE~\cite{maskgae}           & 74.4$_{\pm0.4}$                  & \second{52.6$_{\pm0.6}$} & \best{77.3$_{\pm0.4}$}   & 82.0$_{\pm0.5}$          & 88.6$_{\pm0.6}$          & \second{89.4$_{\pm2.4}$} & \second{82.2$_{\pm0.4}$} \\

& \cellcolor{gray!10}\ours~\circled{6} & \cellcolor{gray!10}74.0$_{\pm0.3}$                  & \cellcolor{gray!10}52.2$_{\pm0.4}$          & \cellcolor{gray!10}\second{77.2$_{\pm0.6}$} & \cellcolor{gray!10}\second{82.3$_{\pm0.7}$} & \cellcolor{gray!10}\second{89.9$_{\pm0.4}$} & \cellcolor{gray!10}88.5$_{\pm1.8}$          & \cellcolor{gray!10}81.5$_{\pm0.6}$          \\

& \cellcolor{gray!10}\ours~\circled{7} & \cellcolor{gray!10}73.8$_{\pm0.5}$                  & \cellcolor{gray!10}52.2$_{\pm0.6}$          & \cellcolor{gray!10}77.0$_{\pm0.2}$          & \cellcolor{gray!10}\second{82.3$_{\pm0.4}$} & \cellcolor{gray!10}\best{90.5$_{\pm0.8}$}   & \cellcolor{gray!10}87.8$_{\pm1.5}$          & \cellcolor{gray!10}81.4$_{\pm0.4}$          \\

& \cellcolor{gray!10}\ours~\circled{8} & \cellcolor{gray!10}73.8$_{\pm0.3}$                  & \cellcolor{gray!10}\best{52.7$_{\pm0.5}$}   & \cellcolor{gray!10}76.1$_{\pm0.4}$          & \cellcolor{gray!10}82.2$_{\pm0.2}$          & \cellcolor{gray!10}89.3$_{\pm0.7}$          & \cellcolor{gray!10}88.4$_{\pm1.0}$          & \cellcolor{gray!10}81.2$_{\pm0.7}$          \\
            \bottomrule
          \end{tabular}
        }
    }
  \end{threeparttable}
\end{table*}

\nosection{Baselines.}
In line with the focus of this work, we benchmark representative GAEs across multiple graph learning tasks, including the vanilla GAE~\cite{gae} and its feature-only variant GAE$_f$, as well as masked GAEs such as MaskGAE~\cite{maskgae}, S2GAE~\cite{s2gae}, GraphMAE~\cite{graphmae}, GraphMAE2~\cite{graphmae2}, and AUG-MAE~\cite{augmae}. 
Among these baselines, GAE, MaskGAE, and S2GAE primarily rely on structure reconstruction, whereas GraphMAE, GraphMAE2, and AUG-MAE adopt feature reconstruction as the learning objective. 
Although GiGaMAE~\cite{gigamae} and SeeGera~\cite{seegera} combine feature and structure reconstruction, they are closer to feature-based GAEs in terms of their training and scalability characteristics.
To provide a more comprehensive comparison, we additionally include several representative GCL baselines, including DGI~\cite{dgi}, GRACE~\cite{grace}, BGRL~\cite{bgrl}, CCA-SSG~\cite{cca_ssg}, GGD~\cite{ggd}, S3GCL~\cite{s3gcl}, and EPAGCL~\cite{epagcl}.
We implement baselines with PyTorch~\cite{pytorch} and PyTorch Geometric~\cite{pyg}.
For feature-based GAEs (GAE$_f$, GraphMAE, GraphMAE2, AUG-MAE, GiGaMAE, and SeeGera), we employ the GAT~\cite{velickovic2018graph} architecture and scaled cosine error (SCE) as the encoder network and learning objective. On the other hand, for structure-based GAEs (GAE, MaskGAE, S2GAE), we utilize the GCN~\cite{kipf2016semi} architecture and binary cross-entropy as the encoder network and learning objective, respectively.

\nosection{Implementation details.}
To align with the baseline implementations, we abstract the contrastive learning principles of GAEs and implement \ours with PyTorch and PyTorch Geometric as well.
According to Table~\ref{tab:cases}, we have seven basic variants of \ours with different contrastive views, and we refer to \ours~\circled{2}\circled{3}\circled{4} as feature-based variants while \ours~\circled{5}\circled{6}\circled{7}\circled{8} as structure-based ones. Since \ours~\circled{2}\circled{3}\circled{4}\circled{5} have implementations proposed in previous works~\cite{grace,graphcl,gae,graphmae,maskgae}, we mainly provide the empirical results of \ours~\circled{6}/\circled{7}/\circled{8} in our experiments.
The hyperparameters of all the baselines were configured according to the experimental settings officially reported by the authors and were then carefully tuned in our experiments to achieve their best results across all tasks. We also tune the hyperparameters of \ours variants for a fair comparison.

\nosection{Hyperparameters of \ours.}
\ours is a stable and general framework, and we use largely consistent hyperparameters across datasets and tasks.
For \ours~\circled{6}/\circled{7}/\circled{8}, we adopt random edge masking as the view augmentation, uniform random negative sampling, and the binary cross-entropy (BCE) objective as the contrastive loss throughout.
For node classification, link prediction, and graph clustering, we use a GCN~\cite{kipf2016semi} encoder, while for graph classification we use a GIN~\cite{xu2018how} encoder.
For heterogeneous node classification, we use GraphSAGE~\cite{hamilton2017inductive} as the encoder.
All encoders have at most two message-passing layers.
For all tasks, we use a two-layer MLP as the decoder.
The encoder hidden dimension is set to 128, and the decoder hidden dimension is set to 32.
We train \ours for up to 500 epochs and apply early stopping based on validation performance.
Unless otherwise specified, we use the Adam optimizer with an initial learning rate of 0.01.

\nosection{Evaluation.}
To provide a comprehensive comparison, we conduct experiments on five graph learning tasks from node, link, subgraph, and graph levels, across homogeneous and heterogeneous graphs, i.e., node classification, link prediction, graph clustering, graph classification, and heterogeneous node classification.
\begin{itemize}
  \item \textbf{Node classification (\red{N}).} Node classification is the most popular graph learning task, with the goal of assigning a class label to each node. In the graph self-supervised learning setting~\cite{maskgae}, the GNN encoder is pretrained based on the pretext tasks to obtain the node embeddings. The final evaluation is done by fitting a linear classifier (i.e., a logistic regression model) on top of the frozen learned embeddings. We adopt the public splits for Cora, CiteSeer, and PubMed, and 8:1:1 training/validation/test splits for the remaining datasets. Classification \textit{accuracy} is employed as the evaluation metric.
  \item \textbf{Link prediction (\blue{L}).} For the link prediction task, the objective is to predict the existence of edges between pairs of nodes. In structure-based GAEs, we directly use the output of the structure decoder to estimate the likelihood of edge existence. In contrast, for feature-based GAEs, the prediction is computed as the dot product of the learned embeddings of the node pairs. For small-scale datasets (Cora, CiteSeer, and PubMed), we follow the 85\%/5\%/10\% training/validation/test splits as recommended by~\cite{gae}. For the large-scale datasets from the Open Graph Benchmark, we use the official public splits provided by the benchmark. We employ the \textit{Area Under the ROC Curve (AUC)} and \textit{Average Precision (AP)} as evaluation metrics for small-scale datasets. For large-scale datasets, we report the Hit rate at rank \(k\) (Hits@\(k\)), which is the standard metric used in the OGB leaderboard.
  \item \textbf{Graph clustering (\orange{C}).} Graph clustering involves grouping nodes in a graph such that nodes in the same group are more similar to each other compared to those in different groups. We perform K-means clustering on the learned embeddings to produce cluster assignments for each node and employ \textit{Normalized Mutual Information (NMI)} as the evaluation metric.
  \item \textbf{Graph classification (\mygreen{G}).} The graph classification task is similar to node classification but differs in that the class labels are assigned to the entire graph. Therefore, we perform \texttt{graph pooling} on the node embeddings to obtain the graph embedding. We follow the evaluation protocol of \cite{graphmae,graphmae2}, which uses a 10-fold cross-validation setting for train-test splits and reports the averaged results. Classification \textit{accuracy} is employed as the evaluation metric.
  \item \textbf{Heterogeneous node classification (\pink{HN}).} We also extend our experiments from common homogeneous graphs to heterogeneous ones, with heterogeneous node classification as the downstream task. In this task, we extend GAEs to heterogeneous graphs by reconstructing each graph view in a manner similar to homogeneous graphs and accumulating the loss from all views to learn the representations of each node type. We adopt the public train/valid/test splits as outlined in \cite{lv2021we} for all datasets. We use classification \textit{accuracy} for the evaluation metric for the target node type in each dataset, i.e., \textit{`author'} (DBLP), \textit{`paper'} (ACM), and \textit{`movie'} (IMDB).
\end{itemize}
For reproduction, we report the averaged results with standard deviations across 10 runs. All experiments are conducted on an NVIDIA RTX 3090 Ti GPU with 24GB memory.

\section{Experimental results}
\label{appendix:exp}

\nosection{Graph clustering.}
Table~\ref{tab:graph_cluster} presents the graph clustering results on seven graph benchmarks. As we employ K-means as the downstream clustering method, we have observed high variances in the NMI results across the majority of datasets. 
Compared with GCL methods, GAEs typically achieve better performance on the graph clustering task. This is attributed to the fact that GAE-style objectives explicitly exploit graph structural signals (e.g., adjacency/edge-induced relationships) during pretraining, which tends to produce embeddings with clearer cluster structure that is well aligned with distance-based clustering. In contrast, standard GCL methods mainly enforce augmentation invariance via instance discrimination; while effective for downstream classification, such invariance does not necessarily translate into well-separated clusters under K-means, especially when the view augmentations introduce nontrivial structural perturbations.
In general, feature-based methods tend to demonstrate higher performance on NMI compared to structure-based methods. However, it is important to note that feature-based methods may suffer from higher memory overheads, particularly on datasets with high-dimensional input features.
For structure-based GAEs, including \ours~\circled{6}/\circled{7}/\circled{8}, they often achieve a more favorable trade-off between performance and scalability compared to feature-based methods. These structure-based approaches leverage the inherent binary graph topology to learn representations, which can be more efficient in terms of memory usage and computational complexity.

\nosection{Graph classification.}
Table~\ref{tab:graph_clas} presents the graph classification results on seven benchmarks. Note that GiGaMAE is not applicable to this task, as it relies on node-level embedding methods (e.g., node2vec~\cite{grover2016node2vec}) to initialize node representations.
We first observe that standard GCL baselines are generally less competitive than GAE-style approaches on these graph-level datasets. A plausible reason is that many GCL pipelines are primarily node-centric and rely on heavy augmentations; when combined with graph-level pooling, such augmentations may distort global semantics and introduce additional variance.
Among GAE-style methods, feature-based GAEs (e.g., GraphMAE) achieve the best overall accuracy, likely because feature reconstruction provides dense supervision and is less sensitive to batching disjoint graphs. That said, our structure-based \ours variants remain competitive, with a relatively small gap to the best feature-based models on most benchmarks. Importantly, \ours~\circled{6}/\circled{7}/\circled{8} do not depend on high-quality node features and can be directly applied when node attributes are missing, noisy, or extremely high-dimensional, where feature-reconstruction methods may become less reliable or incur higher overhead. This robustness makes \ours~\circled{6}/\circled{7}/\circled{8} practical alternatives for graph classification under diverse feature availability and resource constraints.
Overall, these results suggest that while feature reconstruction can be advantageous when rich node features are available, structure-based paired-view learning offers a more generally applicable and scalable solution.

\begin{table}[h]
  \centering
  \small
  \caption{Heterogeneous node classification accuracy (\%) on heterogeneous graph benchmarks.}  
  \label{tab:hetero_node_clas}
   \resizebox{\linewidth}{!}{
  \begin{threeparttable}
    \begin{tabular}{llccc}
      \toprule
                  &      & \textbf{DBLP}                              & \textbf{ACM}             & \textbf{IMDB}              \\ \midrule
      \multirow{4}{*}{\rotatebox{0}{\thead{\textbf{HeteroGNNs}}}} & RGCN~\cite{rgcn}              & 92.07$_{\pm0.50}$                          & 91.75$_{\pm0.35}$        & 65.21$_{\pm0.73}$          \\
      & HAN~\cite{han}               & 92.05$_{\pm0.62}$                          & 90.79$_{\pm0.43}$        & 64.63$_{\pm0.58}$          \\
      & HGT~\cite{hgt}               & 93.49$_{\pm0.25}$                          & 91.15$_{\pm0.71}$        & \second{67.20$_{\pm0.57}$} \\
      & SHGN~\cite{lv2021we}              & 94.20$_{\pm0.31}$                          & 93.35$_{\pm0.45}$        & \best{67.36$_{\pm0.57}$}   \\
      \midrule
      \multirow{6}{*}{\rotatebox{0}{\thead{\textbf{Feature-based}}}} & GAE$_f$~\cite{gae}           & \multicolumn{3}{c}{\texttt{N/A}$^\dagger$}                                                         \\
      & GraphMAE~\cite{graphmae}          & \multicolumn{3}{c}{\texttt{N/A}$^\dagger$}                                                         \\
      & GraphMAE2~\cite{graphmae2}         & \multicolumn{3}{c}{\texttt{N/A}$^\dagger$}                                                         \\
      & AUG-MAE~\cite{augmae}           & \multicolumn{3}{c}{\texttt{N/A}$^\dagger$}                                                         \\
      & GiGaMAE~\cite{gigamae}           & \multicolumn{3}{c}{\texttt{N/A}$^\dagger$}                                                         \\
      & SeeGera~\cite{seegera}           & \multicolumn{3}{c}{\texttt{N/A}$^\dagger$}                                                         \\
      \midrule
      \multirow{6}{*}{\rotatebox{0}{\thead{\textbf{Structure-based}}}} & GAE~\cite{gae}               & 94.1$_{\pm0.3}$                            & 93.3$_{\pm0.5}$          & 65.4$_{\pm0.2}$            \\
      & S2GAE~\cite{s2gae}             & \best{94.8$_{\pm0.4}$ }                    & 93.8$_{\pm0.7}$          & 65.7$_{\pm0.5}$            \\
      & MaskGAE~\cite{maskgae}           & 94.2$_{\pm0.2}$                            & 93.7$_{\pm0.2}$          & 65.9$_{\pm0.7}$            \\

& \cellcolor{gray!10}\ours~\circled{6} & \cellcolor{gray!10}94.4$_{\pm0.2}$                            & \cellcolor{gray!10}93.5$_{\pm0.4}$          & \cellcolor{gray!10}{66.0$_{\pm0.5}$}          \\
& \cellcolor{gray!10}\ours~\circled{7} & \cellcolor{gray!10}\second{94.6$_{\pm0.3}$}                   & \cellcolor{gray!10}\second{94.0$_{\pm0.3}$} & \cellcolor{gray!10}65.2$_{\pm0.6}$            \\
& \cellcolor{gray!10}\ours~\circled{8} & \cellcolor{gray!10}94.3$_{\pm0.2}$                            & \cellcolor{gray!10}\best{94.1$_{\pm0.1}$ }  & \cellcolor{gray!10}{66.3$_{\pm0.4}$ }         \\
\bottomrule
    \end{tabular}
    \begin{tablenotes}
      \item[$\dagger$]\scriptsize Not applicable due to missing node attributes.
    \end{tablenotes}
  \end{threeparttable}
}
\end{table}

\nosection{Heterogeneous node classification.}
To further demonstrate the flexibility and comprehensibility of \ours across various graph learning scenarios, we benchmark experiments of GAEs on three heterogeneous graph datasets, including DBLP, ACM, and IMDB~\cite{lv2021we,hetero2net}.
Table~\ref{tab:hetero_node_clas} presents the experimental results of different GAEs adapted to the heterogeneous node classification task. Several \textit{supervised} heterogeneous GNNs, including HAN~\cite{han}, HGT~\cite{hgt}, RGCN~\cite{rgcn}, and SHGN~\cite{lv2021we} were listed in Table~\ref{tab:hetero_node_clas} for comparison.
Note that we have excluded the comparison of feature-based GAEs due to missing attributes in several node types, which is unavailable to perform feature reconstruction for these datasets. As observed from Table~\ref{tab:hetero_node_clas}, GAEs have shown comparable performance compared to heterogeneous GNNs even without the supervision of class labels.  In particular, \ours~\circled{6}/\circled{7}/\circled{8} achieve generally better performance than other GAEs in most cases, on par with state-of-the-art heterogeneous GNNs.

\begin{figure}[h]
\includegraphics[width=0.58\linewidth]{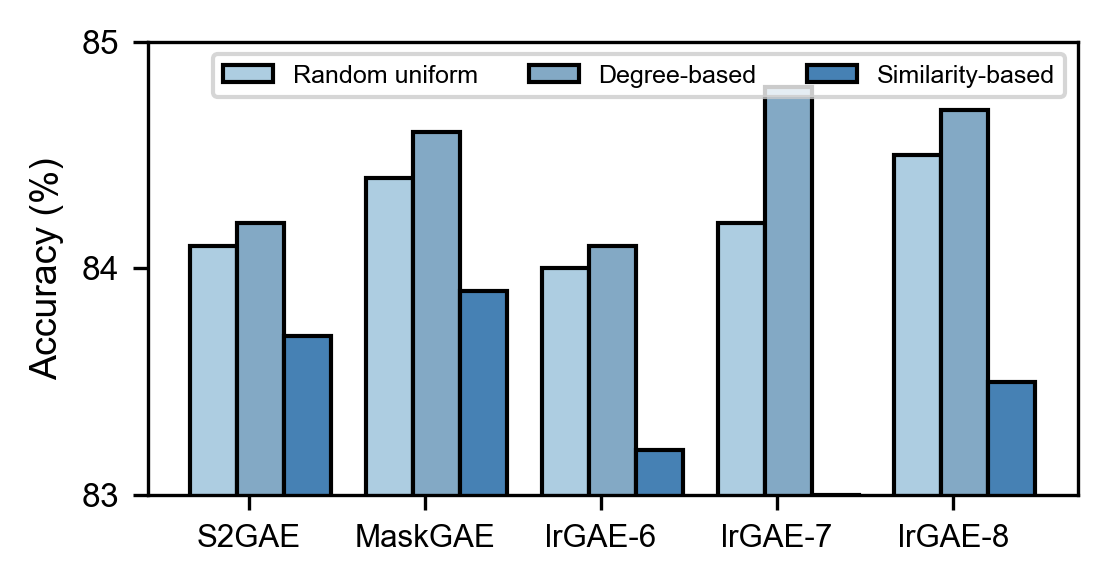}
    \caption{Ablation on the negative sampling strategies.}
    \label{fig:abla_ns}
\end{figure}

\nosection{Ablation on negative samples.}
We explore different negative sampling strategies for GAEs, which are detailed below:

\begin{itemize}
  \item \textbf{Random uniform sampling.}\ Negative edges are sampled uniformly from the set of all possible non-existent edges. This is the simplest approach, but it may not always yield the best results.
  \item \textbf{Degree-based sampling.}\ Negative edges are sampled based on the degree of nodes. The core idea is that nodes with higher degrees are more likely to be involved in many connections, making them more likely to be sampled as negative examples.
  \item \textbf{Similarity-based sampling.}\ Negative edges are sampled based on the similarity or distance between nodes in the graph. For instance, edges between nodes that are far apart are more likely to be sampled as negative. In this case, we use cosine similarity as the distance measure.
\end{itemize}
The ablation results are shown in Figure~\ref{fig:abla_ns}. We observe that random negative sampling generally outperforms other strategies in most cases. Additionally, degree-based sampling shows promise as an effective negative sampling strategy. While similarity-based sampling was expected to improve performance by distinguishing nodes that are more dissimilar, it did not perform as well as the random or degree-based approaches.
The possible reason could be that GAEs necessitate more \textit{hard} negative samples, whereas similarity-based sampling might only furnish \textit{easy} ones.